\documentclass[final,5p,times,twocolumn]{elsarticle}

\usepackage{hyperref}
\usepackage{booktabs}
\usepackage{amsfonts}
\usepackage{algorithm}
\usepackage{algpseudocode}
\algrenewcommand\algorithmiccomment[1]{\hfill{// #1}}

\usepackage{amssymb}
\usepackage{amsmath}
\usepackage{multirow}
\usepackage{multicol}
\usepackage{float}
\usepackage{tcolorbox} 
\usepackage[table,xcdraw]{xcolor}

\journal{Neural Networks}

\begin{document}

\begin{frontmatter}



\title{HVPNet: A Bio-Inspired Network for General Salient and Camouflaged Object Detection}


\author[1]{Jiawei Xu}  
\author[1]{Qiangqiang Zhou\textsuperscript{*}}   
\author[2]{Zhouping Li}   
\author[3]{Yanjiao Shi}   
\author[1]{Yugen Yi}   
\author[1]{Jiacong Yu}   

\affiliation[1]{organization={School of Artificial Intelligence, Jiangxi Normal University}, 
                addressline={Street},
                city={Nanchang},
                postcode={330000},
                state={State},
                country={China}}

\affiliation[2]{organization={Faculty of Business Information, Shanghai Business School}, 
                addressline={Zhongshan West Road},
                city={Shanghai},
                postcode={200235},
                country={China}}

\affiliation[3]{organization={School of Computer Science and Information Engineering, Shanghai Institute of Technology}, 
                addressline={Street},
                city={Shanghai},
                postcode={201418},
                country={China}}

\begin{abstract}
In recent years, most research on multimodal salient object detection (SOD) and camouflaged object detection (COD) typically aims
to improve performance through complex cross-modal feature fusion and decoding structures. However, this approach leads to an
excessively large model parameter scale and often fails to deliver satisfactory detection performance due to structural redundancy. In
contrast, the human visual process is able to efficiently perform salient and camouflaged object identification without such complex
structures. This contrast raises an important question: Can we draw conceptual inspiration from the human visual process to achieve a simpler modeling strategy, and still realize accurate and efficient object detection? To answer this question, we propose HVPNet, a simple yet general bio-inspired computational architecture. Drawing on the multi-layered information integration of the retina as a conceptual metaphor, we designed a Retinal Integration Module (RIM), which effectively integrates multimodal features through a level-specific multi-stage integration strategy. To fully exploit these features, we further design a cortical decoder (CD) that breaks down the decoding process into low- and high-level visual stages, abstracting the hierarchical processing in the human visual cortex. Benefiting from these designs, HVPNet can readily extend to seven tasks across four modalities. Without bells and whistles, it establishes an excellent accuracy-efficiency trade-off across 22 datasets spanning these seven tasks. Our code is available at https://github.com/jiaweiXu1029/HVPNet.
\end{abstract}





\end{frontmatter}



\section{Introduction}
Salient object detection (SOD) and camouflaged object detection (COD) have long been fundamental research topics in the field of computer vision \cite{vscode}. SOD aims to simulate the human visual system to accurately localize the most attention-grabbing objects within an image, whereas COD focuses on identifying objects that are highly blended into their surroundings and thus difficult for human vision to perceive. Both tasks play crucial roles in practical applications such as video surveillance \cite{videosurveillance}, 
intelligent security systems \cite{semanticsegmentation}, object tracking \cite{objectdection}, and video summarization \cite{videosummarization}. With the rapid development of deep learning, research on SOD and COD has achieved remarkable progress. To better accommodate diverse real-world scenarios, researchers have further extended these tasks into various multimodal settings, including RGB SOD, RGB COD, RGB-D SOD, RGB-D COD, RGB-T SOD, as well as video salient object detection (VSOD) and video camouflaged object detection (VCOD).

\begin{figure}[t]
    \centering
    \includegraphics[width=1\columnwidth]{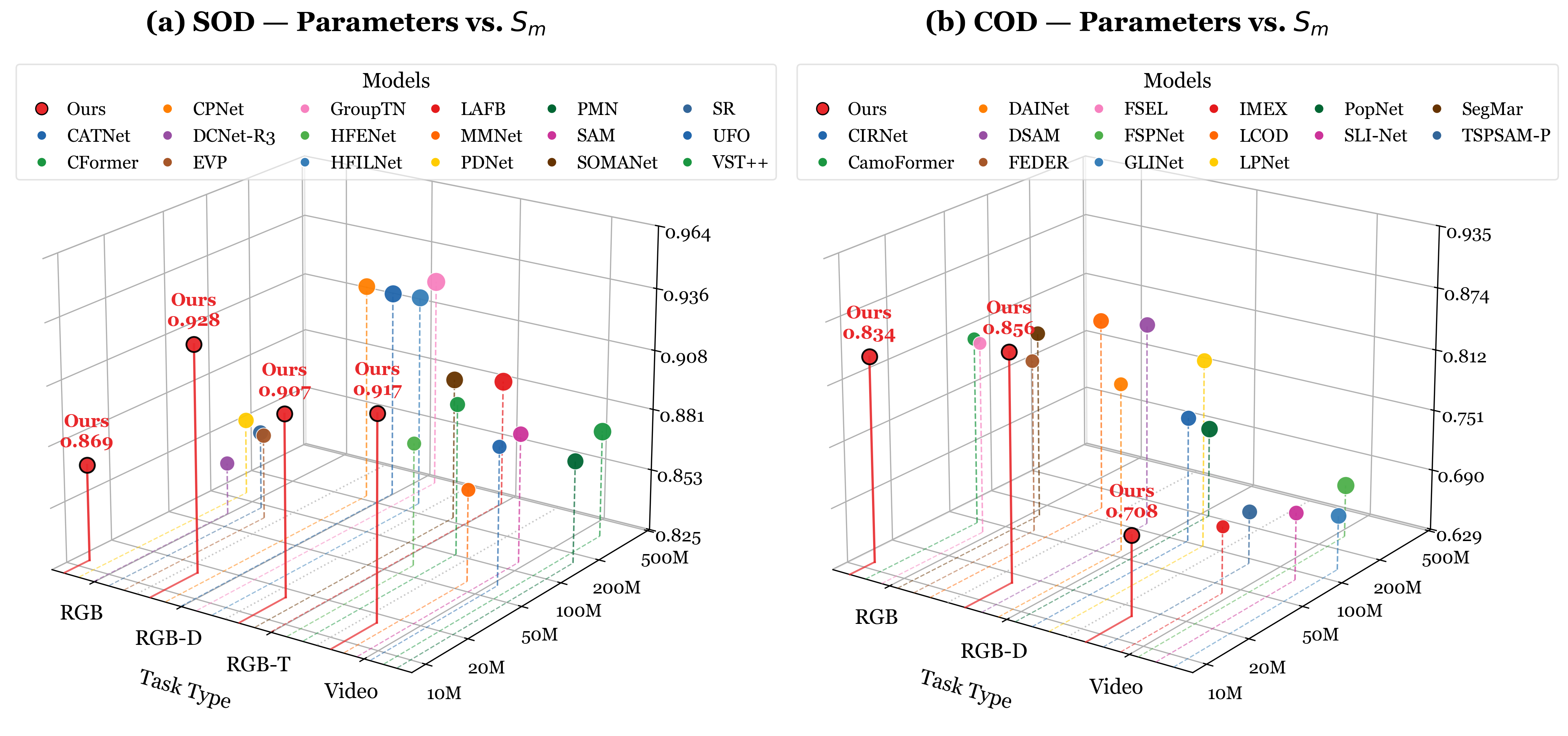}
    \vspace{-3mm}
    \caption{Comparison of parameter counts and  $S_{m}$ scores between HVPNet and previous SOTA methods for SOD and COD. RGB SOD, RGB-D SOD, RGB-T SOD, and VSOD are evaluated on DUT-O~\cite{DUT-O}, STERE~\cite{STERE}, VT821~\cite{VT821}, and SegV2~\cite{Segv2}, respectively; RGB COD and RGB-D COD are evaluated on CAMO~\cite{CAMO}, while VCOD is evaluated on CAD~\cite{CAD}.}
    \label{fig:image1}
    \vspace{-1em}
\end{figure}

In recent years, the rapid development of Vision Transformers has significantly advanced the performance of both SOD and COD. However, as shown in Figure~\ref{fig:image1}, many existing methods~\cite{GroupTransNet, LAFB, camoformer, PopNet, GRNet, HFILNet, EM-Trans, SOMANet, DCNet-R} still rely on a ``heavy extraction, heavy fusion, heavy decoding'' paradigm. These methods often adopt heavyweight backbones, homogeneous fusion modules, and complex decoding structures, leading to substantial computational overhead. For example, CATNet~\cite{CATNet} incorporates attention-based feature enhancement, cross-modal fusion, and cascaded correction decoding, yet requires 262.6M parameters and 314.8G FLOPs. Similarly, CPNet~\cite{CPNET} employs a cross-modal attention fusion module and a progressive decoder, reaching 216.5M parameters and 129.3G FLOPs. This architectural bloat not only significantly increases computational cost but also tends to ignore the inherent differences between multi-level features, causing actual detection performance to encounter a bottleneck. This raises a natural question: are these intricate structures truly necessary? In contrast to such redundant computations, the human visual system relies on concise and highly efficient mechanisms when performing complex perception tasks. This stark comparison prompts us to ask: \textbf{can we draw conceptual inspiration from the human visual process to design a simpler modeling strategy, and still realize accurate and efficient object detection?}

To answer the above question, this paper attempts to explore a network architecture of "lightweight extraction, level-specific simple fusion, hierarchical decoding," aiming to fundamentally rethink the prevailing "heavy extraction, heavy fusion, heavy decoding" paradigm. Specifically, we revisit salient and camouflaged object detection from the perspective of structured multi-level cue coordination, and highlight that selectively organizing semantic and detail representations can be more effective than relying on uniformly heavy processing.
Instead of pursuing marginal gains through increasing architectural complexity, HVPNet is designed to explicitly optimize the trade-off between accuracy and computational efficiency. By abstracting biologically inspired concepts into level-specific lightweight operations, HVPNet achieves competitive SOTA performance while reducing parameters and FLOPs by approximately 90\% compared with many heavyweight counterparts. Specifically, we propose HVPNet, a simple yet effective network for general salient and camouflaged object detection. HVPNet does not claim a direct simulation at the neurophysiological level; rather, it draws conceptual inspiration from the human visual processing mechanism. By abstracting the concepts of the retina's multi-layered integration mechanism and the cortex's hierarchical processing into lightweight neural network operations, HVPNet effectively selectively organizes semantic and detail representations, achieving a substantial reduction in parameters and FLOPs while maintaining highly competitive detection performance.

During the human visual perception process, the retina is not merely a passive phototransduction organ, but a highly intelligent micro-neural network \cite{retina1}. It features a classic three-layered neural circuitry (comprising photoreceptors, bipolar cells, and ganglion cells) capable of multi-scale, multi-level, and multi-granularity spatial information integration \cite{retina2}. The retina performs hierarchical processing ranging from phototransduction and local contrast enhancement to spatial structure extraction and parallel feature channel encoding, ultimately providing structured representations for the cortex. Drawing on this concept of "layered, multi-granularity information integration," we design the Retinal Integration Module (RIM). Unlike existing methods that blindly apply complex, uniform fusion operations across all levels, RIM fully utilizes the multi-scale features extracted by the backbone, applying three fixed and structurally simplified integration strategies specifically tailored for high-level, mid-level, and low-level features. This design conceptually mirrors the functional mapping of the retina's three-stage circuitry, progressively aggregating spatial details and semantic cues into unified multi-level representations. For instance, rather than using heavy global attention everywhere, RIM achieves structured multi-level cue coordination by matching specific operations to the distinct spatial and semantic properties of different feature levels. More importantly, these level-specific simple integration strategies exhibit a high degree of compatibility with lightweight backbones in terms of model capacity. Typically, heavyweight backbones generate massive and complex feature spaces; if simple fusion modules are directly appended after them, it often leads to overfitting due to a mismatch in feature complexity and fusion simplicity. Conversely, the feature representations extracted by lightweight backbones are more compact, which can be highly efficiently integrated by our three-stage simple strategies. This synergistic design effectively avoids the computational redundancy found in traditional methods, elucidating from an architectural perspective the reason why HVPNet can achieve both high efficiency and high performance.

Once visual signals are pre-processed and integrated by the retina, they are transmitted to the brain and processed hierarchically. The primary visual cortex (V1) first decodes basic visual elements such as edges and contours \cite{deocder1}. The information is then routed to higher-level areas (such as V2, V4, and IT) for complex semantic integration \cite{decoder2}. Inspired by this efficient coarse-to-fine processing pipeline, we design a Cortical Decoder (CD). We divide the decoding process into two components: primary visual decoding and high-level visual decoding \cite{cpd}. The primary visual decoding focuses on parsing basic visual features such as edges and contours, explicitly refining detail representations, while the high-level visual decoding handles more complex and abstract information, ensuring the accurate organization of semantic representations. By adopting this hierarchical decoding strategy, we conceptually mirror the brain's coarse-to-fine visual processing pipeline, thereby achieving more efficient image understanding without introducing redundant dense decoders.

In summary, our main contributions are as follows:
\begin{itemize}
    \item We revisit salient and camouflaged object detection from the perspective of structured multi-level cue coordination, and highlight that selectively organizing semantic and detail representations can be more effective than relying on uniformly heavy processing.
    \item We propose HVPNet, a simple yet effective framework for general salient and camouflaged object detection. It translates the conceptual intuition of staged biological vision into an efficient computational architecture.
    \item Within this framework, we design a Retinal Integration Module (RIM) and a Cortical Decoder (CD), which cooperatively enhance multi-level feature interaction and progressive prediction recovery in a compact architecture.
    \item Extensive evaluations on 22 public datasets across seven tasks demonstrate that HVPNet achieves a superior accuracy-efficiency trade-off, maintaining highly competitive performance while significantly reducing computational costs.
\end{itemize}
\section{RELATED WORK}
\subsection{Salient Object Detection}
Early research in salient object detection (SOD) primarily focused on the single RGB modality, with key areas of study including multi-level fusion methods ~\cite{poolnet+,admnet,DMNet-ANN,icon}, and certain recurrent models~\cite{SR,DUT-O,MENet,vst,tp-seg,zhou2026differseg}. With technological advancements, auxiliary modality information such as depth and thermal imaging has become more readily available, leading to a gradual shift in SOD research towards the multi-modal domain. Notably, in RGB-D and RGB-T tasks, the precision of saliency detection has significantly improved.

In the case of RGB-D SOD, most models~\cite{MAGNet,MMNet,HFILNet,EM-Trans} incorporate various attention mechanisms to fuse depth information with RGB features, compensating for the lack of depth data in RGB images. Additionally, some models~\cite{VST++,vscode,CPNET,CATNet} leverage depth priors to further enhance segmentation accuracy, enabling more effective identification of target regions. Similarly, in RGB-T SOD tasks~\cite{LAFB,SOMANet,hfenet,STANet}, attention mechanisms are employed to fully exploit thermal imaging data, complementing RGB information and addressing the details that RGB images alone cannot provide.

For video salient object detection (VSOD), some studies~\cite{SAM(1),PMN,CoSTFormer} focus on extracting spatiotemporal and appearance cues, utilizing both the dynamic temporal changes in video and the static appearance information in images. In recent years, with the maturation of optical flow technology, an increasing number of studies~\cite{ugpl,UFO,MMNet} have integrated optical flow information with appearance details, using motion cues to enhance saliency detection. Optical flow, as a form of modality information, has become a crucial component of modern VSOD tasks. We also view VSOD as a typical multi-modal SOD task, where the fusion and enhancement of different modalities contribute to overall performance improvement.

\begin{figure*}[t]
    \centering
    \includegraphics[width=1\textwidth]{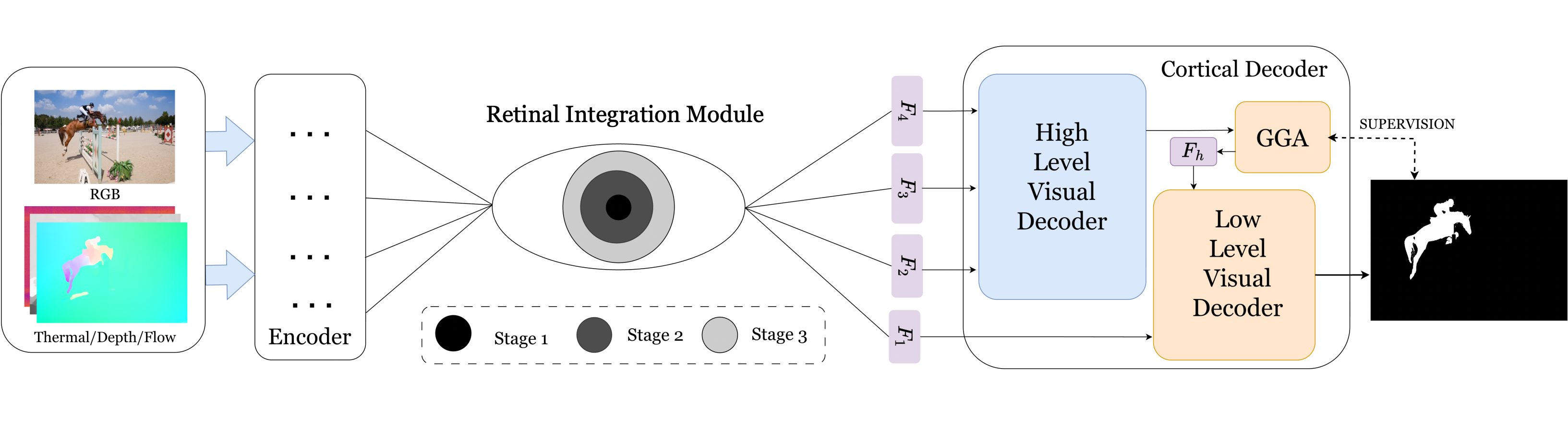}
    \caption{Overall architecture of the proposed HVPNet for general SOD and COD tasks. It consists of three stages: feature extraction, fusion, and cortical decoding (high-level visual decoding, low-level visual decoding).}
    \label{fig:image2}
\end{figure*}
\subsection{Camouflaged Object Detection}
Compared with Salient Object Detection (SOD), Camouflaged Object Detection (COD) emerged relatively later. However, with the increasing demand for this task and its expanding application scenarios, the field has gradually attracted significant research interest. In RGB COD, existing methods~\cite{CAMO,segMar,FSEL,CDP} primarily fall into three representative categories: multi-task learning approaches~\cite{CDP,vscode}, which enhance camouflaged object perception by jointly learning related tasks; multi-input approaches~\cite{DSAM,CIPNet}, which exploit multi-scale, multi-context, or multi-view inputs to enrich feature representations; and refinement approaches~\cite{NC4K,LCOD}, which further improve detection accuracy through boundary refinement and camouflaged region enhancement.

With the growing availability of auxiliary modalities, RGB-D COD was first introduced in~\cite{PopNet}, where depth inference models~\cite{DAINet,CIPNet} were adapted for camouflaged object segmentation, providing additional structural cues for understanding camouflaged scenes. Subsequently, increasing research efforts have focused on leveraging the complementary nature of RGB and depth modalities to develop more robust and efficient RGB-D COD frameworks capable of handling complex backgrounds and highly camouflaged scenarios.

For Video Camouflaged Object Detection (VCOD), similar to the paradigm of Video Salient Object Detection (VSOD), motion cues play a crucial role in improving detection performance. Recent studies~\cite{FSPNet,FEDER,HGINet} commonly integrate optical flow information with appearance features to extract key motion consistency and background dynamics, enabling better discrimination between camouflaged objects and their surroundings. Based on these observations, we also regard VCOD as a typical multi-modal COD task.
\subsection{Bio-inspired and Cognitive Methodologies}
Object detection problems are closely aligned with the mechanisms underlying human visual perception and cognitive reasoning. Consequently, recent studies have increasingly explored brain-inspired methodologies to alleviate visual ambiguities in both salient object detection (SOD) and camouflaged object detection (COD).

From the perspective of perceptual organization, early SOD works~\cite{itti,twostream,salgan} simulated biological center-surround receptive fields or the dual visual pathways of the human brain to separately process appearance and spatial information. To address visual ambiguities arising from indistinct foreground–background appearances, figure-ground assignment mechanisms~\cite{exploring} have been investigated and shown to be effective in improving perceptual grouping. Building upon these cognitive principles, subsequent COD research further incorporated human-like structural priors to tackle extreme camouflage scenarios. For instance, models such as~\cite{mirrornet,DtcNet} employ dual-stream biological attack mechanisms or texton-coherence statistics to enhance the discrimination of ambiguous targets from complex textures.

Beyond static structural cues, another research direction draws inspiration from the dynamic and probabilistic nature of human cognition. Motivated by eye-movement experiments, BiCOD~\cite{bicod} derives cognitive laws to guide attention-based feature extraction. To mimic the sequential visual search behavior of humans, recurrent glimpse-based decoders~\cite{2022recurrent} have been introduced to iteratively refine attention over potential object regions. Furthermore, addressing the intrinsic uncertainty of camouflage, UGTR~\cite{uncertainty} models human reasoning under uncertain conditions through probabilistic representations.

Collectively, these studies demonstrate that incorporating biological principles and cognitive paradigms can enhance detection robustness. They provide conceptual motivation for our lightweight, brain-inspired architectural design.

\section{PROPOSED METHOD}
The architecture of HVPNet, illustrated in Figure~\hyperref[fig:image2]{\textcolor{red}{2}}, consists of an encoder, a retinal integration module (RIM), and a cortical decoder (CD). Initially, the encoder independently extracts rgb features \(
\{ F_1^{r}, F_2^{r}, F_3^{r}, F_4^{r} \}
\) and auxiliary features \(
\{ F_1^{x}, F_2^{x}, F_3^{x}, F_4^{x} \},
\) from the RGB image and the auxiliary images (e.g., depth map, thermal map, or optical flow). These multi-level features are then progressively integrated through the RIM to enhance complementary information. ly, the integrated representations are fed into the CD, which hierarchically refines the features to generate the final prediction map.
\begin{figure*}[t]
    \centering
    \includegraphics[width=1\textwidth]{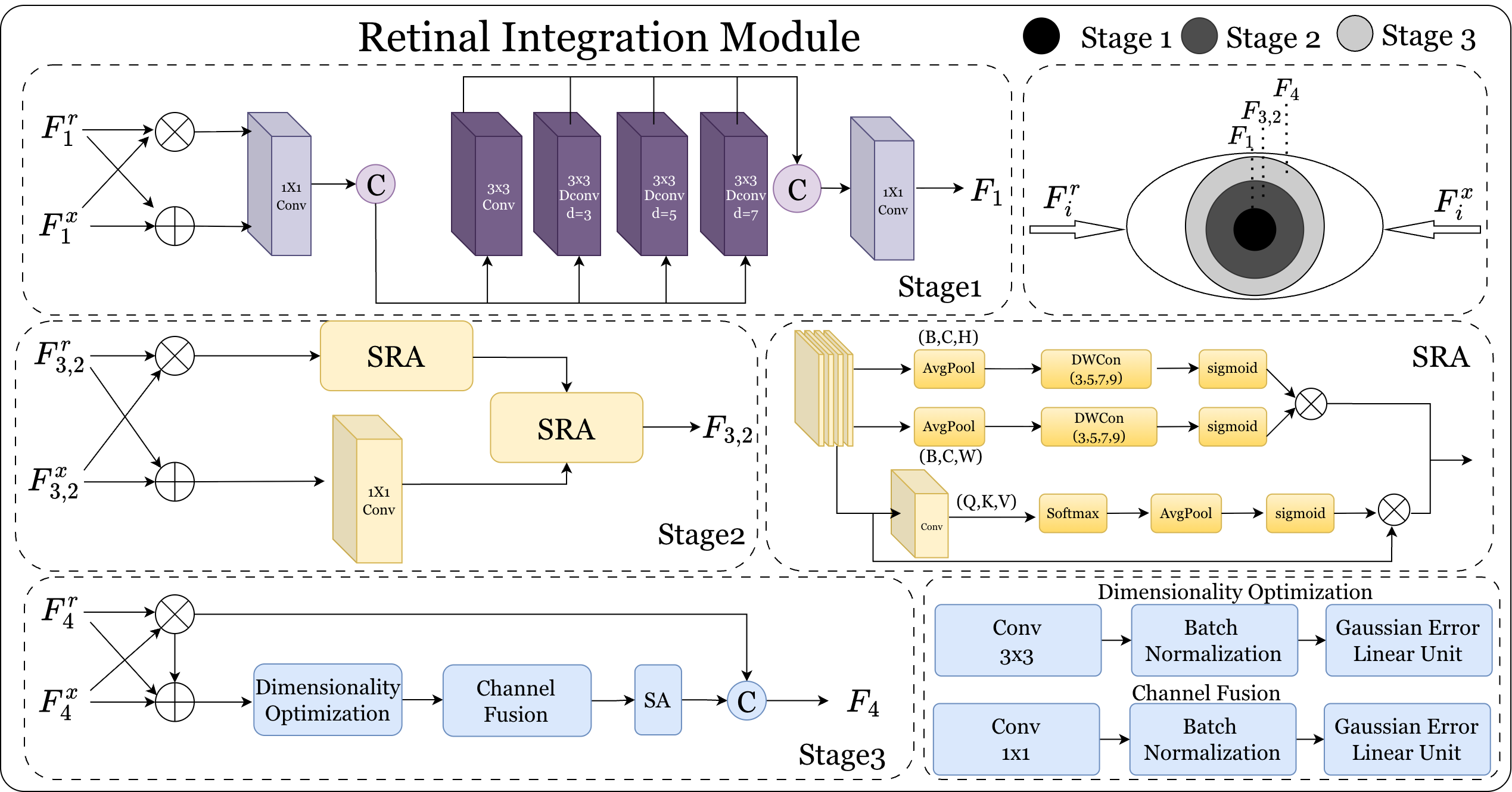}
    \caption{Illustration of the retinal integration module (RIM). We employ three distinct stages to address the specificities of features at different levels. This process ultimately yields cross-modal integrated features \(\{F_1, F_2, F_3, F_4\}\) across various levels.}
    \label{fig:image3}
\end{figure*}

\subsection{Retinal Integration Module}
Drawing conceptual inspiration from the classic three-layered neural circuitry of the retina, illustrated in Figure~\hyperref[fig:image3]{\textcolor{red}{3}}, we propose a Retinal Integration Module (RIM) that eschews uniform, computation-heavy fusion. Instead, it uses three stages (stage~1, stage~2, and stage~3) to perform tailored, level-specific integration on the feature sets \(
\{ F_1^{r}, F_2^{r}, F_3^{r}, F_4^{r} \}
\) and \(
\{ F_1^{x}, F_2^{x}, F_3^{x}, F_4^{x} \},
\) which are obtained from the encoder.

 \noindent \textbf{Stage~1.} To enhance the model’s ability to perceive local structures and contours, 
we focus on integrating low-level features that emphasize edge information in 
stage~1. Low-level edge features play a crucial structural role in 
visual tasks, providing more accurate shape priors for subsequent semantic 
modeling. Therefore, we first perform element-wise addition and multiplication 
between the low-level features \( F_1^{r} \) and \( F_1^{x} \) from different 
modalities to explicitly model their complementarity in edge responses. The 
results are then integrated through a 1\(\times\)1 convolution to produce the 
initial fused feature \( F_1^{f} \).

To further capture local contextual information and multi-scale edge 
structures, \( F_1^{f} \) is sequentially processed by a 3\(\times\)3 
convolution for local detail extraction, followed by three 3\(\times\)3 
dilated convolution layers with dilation rates \( d = 3, 5, 7 \). These 
dilated convolutions expand the receptive field without introducing 
additional computational cost, enabling the network to effectively extract 
contour structures at multiple scales. ly, the outputs of all dilated 
convolutions are integrated using a 1\(\times\)1 convolution, generating the 
integrated output feature \( F_1 \) for stage~1.
The detailed computation is formulated as follows:
\begin{equation}
F_1^{f} = \phi(F_1^{r} \oplus F_1^{x}) + \phi(F_1^{r} \otimes F_1^{x}),
\end{equation}
\begin{equation}
F_1 = \phi\!\left( \psi(F_1^{f}) \oplus \sum_{i \in \{3,5,7\}} D_i(F_1^{f}) \right),
\end{equation}
where \( \phi(\cdot) \) denotes a 1\(\times\)1 convolution, 
\( \psi(\cdot) \) denotes a 3\(\times\)3 convolution,
\( D_i(\cdot) \) denotes a 3\(\times\)3 dilated convolution with dilation rate \( i \),
\( \otimes \) denotes element-wise multiplication, 
and \( \oplus \) denotes element-wise addition.

 \noindent \textbf{Stage~2.} Mid-level features typically contain more semantic information and are better at capturing the structure and semantic relationships within an image. Therefore, in Stage 2, we focus on the selective extraction of relevant and prominent features between the two modalities. Two branches are used: one branch performs element-wise multiplication on \( F_i^{r} \) and \( F_i^{x} \) (i = 2, 3), and the result is fed into the Selective Region Attention (SRA) module to enhance important features, producing \( F_i^{f} \). The motivation behind this step is to emphasize the correlation between the modalities through multiplication and highlight the most important features using the SRA module. The second branch processes the features through a 1×1 convolution, adds the result to \( F_i^{f} \), and then passes the combined output to the SRA module for further enhancement. The goal of this process is to not only preserve valid information during feature fusion but also strengthen the representation of key regions.

Within the SRA module, the input feature \( X \in \mathbb{R}^{B \times C \times H \times W} \) is processed through three separate branches: the spatial width attention branch, the spatial length attention branch, and the channel attention branch. This design allows us to precisely adjust the weights of features across different dimensions. Initially, both the spatial width and spatial length attention branches compute the global information of the feature map through average pooling, followed by convolutions with four different dilation rates \( \{ d = 3, 5, 7, 9 \} \) to generate corresponding attention maps. With this design, the SRA module effectively learns feature information extracted from different scales and processes it using a Sigmoid activation function. Finally, these attention maps are fused and combined with the attention map produced by the channel attention branch to generate the final output \( F_i \). The detailed formulas are given below:
\begin{equation}
SRA(X) = A_{spatial} + A_{channel},
\end{equation}
\begin{equation}
F_i = {SRA}\left( \text{conv}( F_i^{r} \oplus F_i^{x}) \right) \oplus {SRA}\left( (F_i^{r} \otimes F_i^{x}) \right),
\end{equation}
where \( \otimes \) denotes element-wise multiplication, and \( \oplus \) denotes the addition operation.

\noindent \textbf{Stage~3.} High-level features typically contain richer semantic information, making them crucial for image understanding and reasoning. To maximize the expressive power of these features, the goal of Stage 3 is not only to integrate multimodal information but also to enhance the network's ability to perceive global semantic information. First, \( F_4^{r} \) and \( F_4^{x} \) are processed through element-wise multiplication on the feature maps. Then, the result of the element-wise multiplication is fused with the original feature maps along the channel dimension. The fused feature map is further processed by a dimensionality reduction optimization module and a channel fusion module. Specifically, the dimensionality reduction optimization module consists of a 3x3 convolution layer, Batch Normalization (BN), and a Gaussian Error Linear Unit (GELU), while the channel fusion module replaces the 3x3 convolution layer in the dimensionality reduction module with a 1x1 convolution layer to better facilitate information fusion and enhance the network's expressive power. Subsequently, a spatial attention mechanism is employed to achieve global feature perception of the feature map, helping the model better understand the overall context of the image. Finally, the resulting feature map is fused with global semantic information. The specific formula is as follows:
\begin{equation}
DO(X) = \sigma(conv3* 3 (BN(X))),
\end{equation}
\begin{equation}
CF(X) = \sigma(conv1 * 1(BN(X))),
\end{equation}
\begin{equation}
F_4 = (F_i^{4} \otimes F_4^{x}) \oplus SA(CF(DO((F_4^{r} \otimes F_4^{x}) \oplus (F_4^{r} \oplus F_4^{x})))),
\end{equation}
where  DO represents dimensionality reduction optimization, CF represents channel fusion, and \(\sigma \) and BN represent the GELU activation function and batch normalization layer, respectively.

\begin{figure}[t]
    \centering
    \includegraphics[width=\columnwidth]{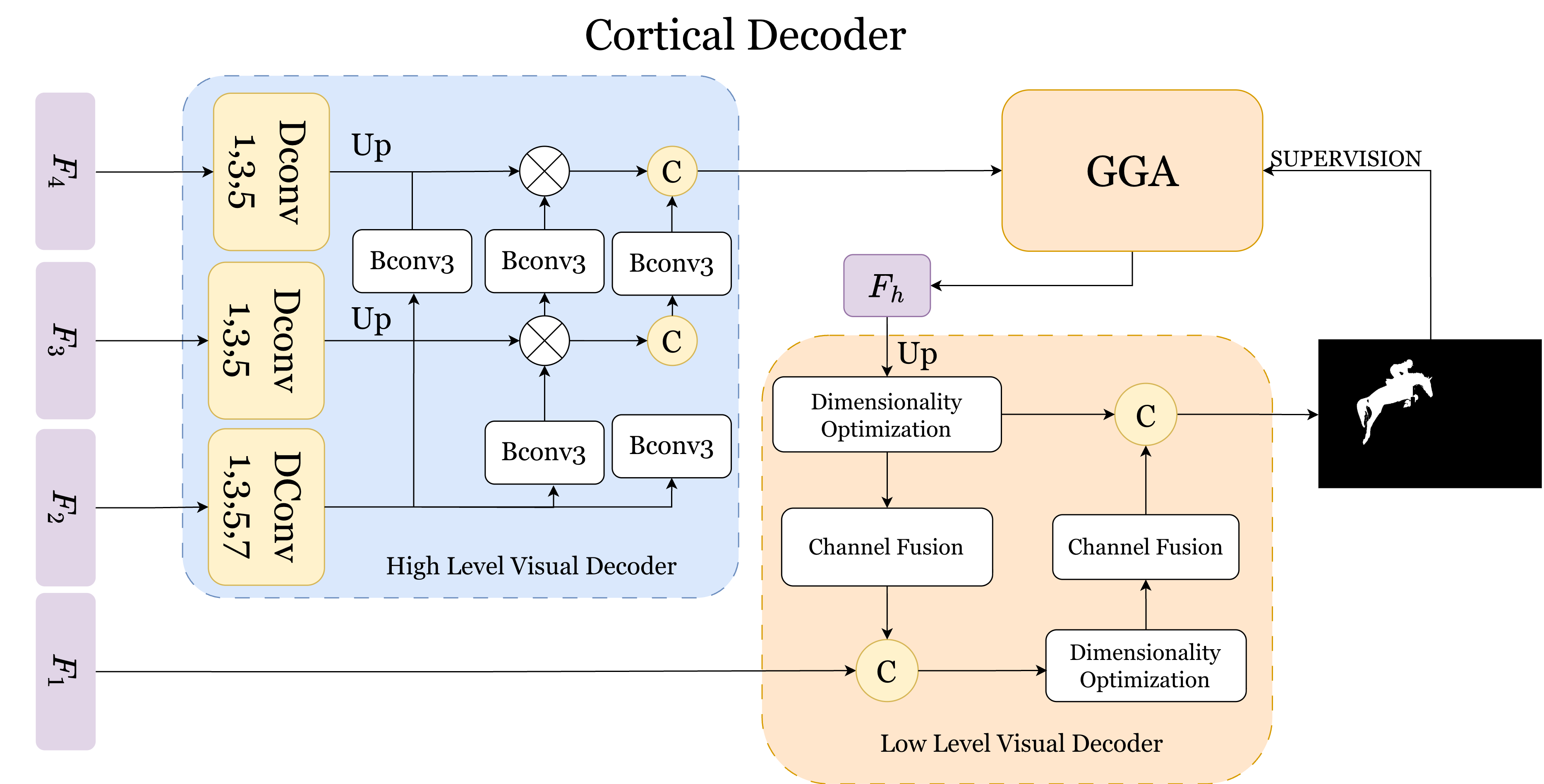}
    \caption{Illustration of the cortical decoder (CD) module.}
    \label{fig:image4}
\end{figure}
\subsection{Cortical Decoder}
The cerebral cortex processes visual information hierarchically. Drawing conceptual inspiration from this layered paradigm, we propose a cortical decoder (CD) that includes both a low-level visual decoder (LLVD) and a high-level visual decoder (HLVD). Rather than employing standard monolithic decoding layers, this design explicitly separates semantic abstraction and detail refinement to provide an efficient and generalized decoding process: HLVD captures universal semantics, while LLVD refines details, allowing the same structure to generate accurate predictions for both SOD and COD across modalities with significantly reduced computational overhead.

\subsubsection{High-Level Visual Decoder}
Higher visual areas encode abstract semantics at lower resolutions but lack fine spatial details, necessitating modulation to guide lower areas. Biologically, areas like V4 and IT perform hierarchical integration, fusing multi-level abstractions to capture global context, which can be modeled as a fusion process with spatial smoothing to enhance semantic consistency. To emulate this, we divide the features \(\{F_1, F_2, F_3, F_4\}\) into high-level \(\{F_2, F_3, F_4\}\) and low-level (\(F_1\)) groups. HLVD uses dilated convolutions for expanded receptive fields, depthwise separable convolutions for efficient channel-wise processing, and upsampling for resolution alignment. This generates an original attention map \(S\):

\begin{equation}
    S = \text{Fuse}(F_2, F_3^{\uparrow}, F_4^{\uparrow}),
\end{equation}
where \(\text{Fuse}(\cdot)\) denotes convolutional fusion with upsampling (\(\uparrow\)) to align spatial resolutions.

\subsubsection{Gaussian Guide Attention}
To enhance modulation, we introduce Gaussian Guide Attention (GGA), which applies a 31×31 Gaussian filter (\(\sigma=4\)) for spatial smoothing, followed by max-min normalization \(\mathcal{N}\). This simulates cortical spatial integration and receptive field modulation: the Gaussian kernel models smooth blending of semantic cues, while the max operation integrates local details with global context, substituting the biological tuning of higher areas. The guided feature is:
\begin{equation}
     F_h = \max\left(\mathcal{N}\left(G_\sigma * S\right),\ S\right),
\end{equation}
where \(G_\sigma\) is a 2D Gaussian kernel. This design rationalizes high-level simulation by providing efficient, bio-aligned modulation. The core process of GGA is detailed in Algorithm 1.
\begin{algorithm}[t]
\caption{Core process of GGA}
\begin{algorithmic}[1]
\State \textbf{Input:} Attention map $A$ and feature map $X$, both with shape $(\text{batch\_size}, \text{channels}, \text{height}, \text{width})$
\State \textbf{Output:} Refined feature map $X$ after global Gaussian attention
\State Initialize a Gaussian kernel
\State Convert $K$ to a 4D tensor and set it as a learnable parameter
\State \textbf{Forward process:}
\State $S \gets F.conv2d(A, K, \text{padding}=15)$ \Comment{Apply Gaussian smoothing on $A$}
\State $S \gets \text{min\_max\_norm}(S)$ \Comment{Normalize the smoothed attention map}
\State $X \gets X \cdot \max(S, A)$ \Comment{Fuse normalized attention with input feature map}
\State \textbf{Return:} $X$
\end{algorithmic}
\end{algorithm}

\subsubsection{Low-Level Visual Decoder}
Early visual areas like V1 retain rich spatial details but lack semantic depth, relying on top-down modulation from higher areas for refinement. Specifically, we use \(F_h\) to optimize \(F_1\), ensuring that high-level semantics guide the refinement of low-level details without over-smoothing the abstract features. Biologically, this involves hierarchical fusion where semantic cues suppress noise and enhance details, modeled as iterative refinement with attention-like mechanisms. To substitute this, the LLVD refines \(F_h\) through the DO and CF to boost semantics (simulating top-down input), then fuses it with \(F_1\) for detail integration. Another further refines the combined feature, producing a prediction map with clear boundaries and rich semantics. This progressive refinement mimics V1's role in boundary enhancement under higher cortical guidance. The output
is the predicted map P.
\subsection{Loss Function}
In this study, we use a composite loss function that combines Cross-Entropy Loss (CE)~\cite{CELoss} and Intersection over Union Loss (IOU)~\cite{IOU} to optimize the model's performance:
\begin{equation}
CE(\hat{y},y)= -\frac{1}{N} \sum_{i=1}^{N} \left[ y \log(\hat{y}) + (1 - y) \log(1 - \hat{y}) \right],
\end{equation}
\begin{equation}
IOU \, Loss = 1 - IOU(A, B).
\end{equation}

We designed four prediction outputs, each of which simultaneously calculates both the CE loss and the IOU loss. Specifically, the loss for the  i\text{-th }prediction is given by the following formula:
\begin{equation}
{Loss}_i = CE(\hat{y}, y) + IOU \, Loss(\hat{y}_i, y),
\end{equation}
the total loss function is the sum of the four losses:
\begin{equation}
{Total Loss} = {Loss}_1 +{Loss}_2 +{Loss}_3 + {Loss}_4.
\end{equation}

\begin{table*}[ht]
    \centering 
    \caption{Quantitative comparison of our HVPNet against other SOTA \textbf{RGB SOD} methods. Bolded results are the best.}
    \setlength{\tabcolsep}{0.3mm} 
\resizebox{\linewidth}{!}{
\begin{tabular}{c|cc|cccc|cccc|cccc|cccc}
\toprule \multirow{2}{*}{Methods} & \multirow{2}{*}{Params(M)$\downarrow$} & \multirow{2}{*}{FLOPs(G)$\downarrow$}  & \multicolumn{4}{c|}{DUT-O \cite{DUT-O}} & \multicolumn{4}{c|}{DUTS \cite{DUTS}}  & \multicolumn{4}{c}{HKU-IS \cite{HKU-IS}}& \multicolumn{4}{c}{ECSSD \cite{ECSSD}} \\
&  & & $E_m \uparrow$ & $S_m \uparrow$ & $F_m \uparrow$  & $\mathcal{M} \downarrow$ & $E_m \uparrow$ & $S_m \uparrow$ & $F_m \uparrow$  & $\mathcal{M} \downarrow$& $E_m \uparrow$& $S_m \uparrow$ & $F_m \uparrow$& $\mathcal{M} \downarrow$& $E_m \uparrow$ & $S_m \uparrow$ & $F_m \uparrow$ & $\mathcal{M} \downarrow$\\ \hline 
\rowcolor{yellow!20} \multicolumn{19}{c}{Lightweight Methods} \\
PoolNet-M+(TPAMI'23)\cite{poolnet+}& 3.0 & .928  &.873 &.830 &.765&\textbf{.056} & .849 & .868  &.852 &.046 & .946 & .899 & .905&.037& - & .909 & .912 &.048 \\
ADMNet+(TIM'24)\cite{admnet}& 0.8 & 3.3  &.869 &.826 &.763&.058 & .893 & .835  &.813 &.052 & .946 & .906 & .901&.036& .933 & .900 & .909& .049 \\
DMNet-ANN(AAAI'25)\cite{DMNet-ANN}& 2.6 & 43.5  &- &.822 &.751&.057 & - & \textbf{.879}  &.848 &.042 &- & \textbf{.925} & \textbf{.920}&\textbf{.031}& - & .920 & .923&\textbf{.039}\\

 \textbf{HVPNet(-)} & 2.8 & 1.8& \textbf{.881} & \textbf{.830} &\textbf{.777} &.057& \textbf{.921} & .870 & \textbf{.852} &\textbf{.041} &\textbf{ .952} & .905 & .913 &.034& \textbf{.946} & .911 & \textbf{.923}&.041 \\

 \rowcolor{red!10} \multicolumn{19}{c}{Accurate Methods} \\
 ICON-R(TAPMI'23)\cite{icon} & 33.0 & 20.9 & .884 & .845& .799&.057 & .931 & .890 & .876&.037 & .960 & .920 & .931&.029 & .960 & .929 & .943 &.032\\
 MENet(CVPR'23)\cite{MENet}& 27.8 & 94.7 & .879 & .850& .792&.045 & .943 & .905 & .895&.028 & .965 & .927 & .939&.023& .954 & .928 & .955&.031\\
 EVP(CVPR'23)\cite{EVP} & 64.5 & - & .902 & .864&.822& .047 & .943 & .905 & .895&.028& .971 & .935 & .945&.024& .965 & .936 & .949&.029\\
SR(TIM'24)\cite{SR}& 90.7 & 26.2 & .894 & .861& .807&.043 & .952 & .911 & .902 &.027 & .969 & .931 & .941&.024& .965 & .936 & .948&.027\\
VST-T++(TAPMI'24)\cite{VST++} & 54.6 & - & .892 & .853& .804 &.053& .943 & .901 & .887&.033 & .968 & .930 & .939&.026& .968 & .937 & .949&.029\\
DCNet-R(PR'25)\cite{DCNet-R}& 356.3 & - & .876 & .849 & \textbf{.827}&.053 & .927 & .896 & .899&.035 & .954 & .924 & .942&.027& .945 & .924 & .949&.034\\
PDNet(EAAI'25)\cite{PDNet}& 116.0 & 18.2 & .900 & .860 & .814&.049& .950 & .912 & .900&.027 & .972 & .933 & .944&.023& \textbf{.973} & \textbf{.943} & .956&.024\\
 HVPNet &\textbf{11.7} &\textbf{8.1} &\textbf{.905} & \textbf{.869} &\textbf{.827}&\textbf{.043}  & \textbf{.950} & \textbf{.913} & \textbf{.905}&\textbf{.026} & \textbf{.972} &\textbf{.935} & \textbf{.946}&\textbf{.021} & .969 & .940 & \textbf{.957}&\textbf{.024}\\
\hline
\end{tabular}
\label{tab:1}
}
\vspace{-5mm}
\end{table*}
\begin{table*}[ht]
    \centering 
    \caption{Quantitative comparison of our HVPNet against other SOTA \textbf{RGB-D SOD} methods. Bolded results are the best.}
    \setlength{\tabcolsep}{0.3mm} 
\resizebox{\linewidth}{!}{
\begin{tabular}{c|cc|cccc|cccc|cccc|cccc}
\toprule \multirow{2}{*}{Methods} & \multirow{2}{*}{Params(M) $\downarrow$} & \multirow{2}{*}{FLOPs(G) $\downarrow$}  & \multicolumn{4}{c|}{NLPR \cite{NLPR}} & \multicolumn{4}{c|}{NJUD \cite{NJU2K}} & \multicolumn{4}{c}{DUT-Depth \cite{DUT-Depth}}& \multicolumn{4}{c}{STERE \cite{STERE}}  \\
&  &  & $E_{\varepsilon}^{\max } \uparrow$ & $S_\alpha \uparrow$ & $F_\beta^{\max } \uparrow$ & $\mathcal{M} \downarrow$ & $E_{\xi}^{\max } \uparrow$ & $S_\alpha \uparrow$ & $F_\beta^{\text {max }} \uparrow$ & $\mathcal{M} \downarrow$ & $E_{\xi}^{\max } \uparrow$ & $S_\alpha \uparrow$ & $F_\beta^{\text {max }} \uparrow$ & $\mathcal{M} \downarrow$& $E_{\varepsilon}^{\max } \uparrow$ & $S_\alpha \uparrow$ & $F_\beta^{\max } \uparrow$ & $\mathcal{M} \downarrow$\\ \hline 
\rowcolor{yellow!20} \multicolumn{19}{c}{Lightweight Methods}  \\
 LSNet(TIP'23)\cite{lsnet} & 4.6 & 1.2 & .961 & .919 & .910 & .025 & .950  & .911 & .914& .925 & .871 & .864 & .054  & .039 & .882 & .838 & .821 & .073 \\
 MAGNet(KBS'24)\cite{MAGNet} & 5.2 & 2.5 & .958 & .916 & .906 & .025 & .946 & .907 & .908  & .038  & .950 & .914 & .920 & .035 & .938 & .888 & .888 & .045 \\
 AirSOD(TCSVT'24)\cite{airsod}& 2.4 & 0.9  & .963 & .924 & \textbf{.923} & .023 & .944 & .908  & .918 & .039 & .946 & .891  & .920 & .048 & .939 & .895 & .900 & .043\\
HVPNet(-) & 3.6 & 3.4 & \textbf{.965} & \textbf{.927} & .920 & \textbf{.021} & \textbf{.953} & \textbf{.919} & \textbf{.922} & \textbf{.033} & \textbf{.958} & \textbf{.928} & \textbf{.935} &\textbf{.028} & \textbf{.946} & \textbf{.903} & \textbf{.903} & \textbf{.038}\\

 \rowcolor{red!10} \multicolumn{19}{c}{Accurate Methods} \\
 PopNet(ICCV'23)\cite{PopNet} & 198.9 & 108.0 & - & - & - & -& .952 & .924 & .936 & .030 & - & - & - & -& .947 & .917 & .924 & .033 \\
 MMNet(TIP'24)\cite{MMNet} & 80.0 & 24.5  & .965 & .934 & .923 & .023 & .959 & .932 & .934 & .031 & .969 & .948 & .952 & .023& .955 & .922 & .919 & .035 \\
 STANet(NEUCOM'24)\cite{STANet} & 85.0 & 24.5 & .968 & .934 & .926 & .021 & .954 & .924 & .927 & .034 & - & - & - & -& .952 & .915 & .913 & .037  \\
 EM-Trans(TNNLS'24) \cite{EM-Trans}& - & - & .970 & .941 & .932 & .018 & .960 & .931 & .934 & .028 & - & - & - & -& .956 & .926 & .924 & .028 \\
HFILNet(TOMM'24)\cite{HFILNet}& 246.5 & 242.2 & .969 & .943 & .915 & .016 & .939 & .936 & .931 & .025 & .969 & .951 & .951 & .020 & .934 & .922 & .903 & .029\\
GroupTransNet(Neurocom'24)\cite{GroupTransNet}& 431.6 & - & .961 & .928 & .908 & .019 & .926 & .922 & .921 & .028 & .958 & .935 & .939 & .024 & .928 & .908 & .895 & .032\\
CATNet(TMM'24)\cite{CATNet}& 262.5 & 341.8 & .966 & .940 & .923 & .018 & .956 & .932 & .928 & .026 & .971 & .953 & .950 & .020& .952 & .921 & .905 & .030 \\
 CPNet(IJCV'25)\cite{CPNET}& 216.5 & 129.3 & .969 & .940 & .918 & .016 & .934 & .934 & .920 & .025 & .971 & .950 & .954 & .019& .934 & .920 & .903 & .029  \\
 MambaSOD(Neurocom'25)\cite{mambasod}& 78.9 & - & \textbf{.973} & .941 & .934 & .017 & .963 & .934 & .937 & .027 & .967 & .937 & .935 & .019 & .955 & .924 & .920 & .031 \\
 HVPNet & \textbf{16.6} & \textbf{11.5} & .971 & \textbf{.942} & \textbf{.934} & \textbf{.016} & \textbf{.96}4 & \textbf{.935} & \textbf{.940} & \textbf{.024} & \textbf{.974} & \textbf{.951} & \textbf{.960 }& \textbf{.017} & \textbf{.962} & \textbf{.928} & \textbf{.929} &\textbf{.026} \\
\hline
\end{tabular}
\label{tab:2}
}
\end{table*}

\section{EXPERIMENTS}
\subsection{Datasets}
\label{chap:4.1}
For \textbf{RGB SOD}, we evaluated HVPNet on three commonly used benchmark datasets: DUTS \cite{DUTS}, DUT-O \cite{DUT-O}, and HKU-IS \cite{HKU-IS}.
For \textbf{RGB-D SOD}, we evaluated HVPNet on three benchmark datasets: NJUD \cite{NJU2K}, NLPR \cite{NLPR}, and STERE \cite{STERE}.
For \textbf{RGB-T SOD}, we evaluated HVPNet on three benchmark datasets: VT821 \cite{VT821}, VT1000 \cite{VT1000}, and VT5000 \cite{VT5000}.
For \textbf{VSOD}, we evaluated HVPNet on two widely used benchmark datasets: FBMS \cite{FBMS} and SegV2 \cite{Segv2}.
For \textbf{RGB COD} and \textbf{RGB-D COD}, we evaluated HVPNet on the same three datasets: COD10K \cite{COD10K}, NC4K \cite{NC4K}, and CAMO \cite{CAMO}.
For \textbf{VCOD}, we evaluated HVPNet on one widely adopted benchmark dataset: CAD \cite{CAD}. 
\subsection{Evaluation metrics}
To ensure a consistent evaluation across all SOD and COD tasks, we employed four saliency evaluation metrics: Structure Measure ($S_{m}$)~\cite{SA}, Mean Absolute Error ($M$)~\cite{Mae}, Maximum F-measure ($F_{m}$)~\cite{Fmeasure}, and Maximum Enhanced Alignment Measure ($E_{m}$)~\cite{EMAX}. Additionally, to assess the model’s computational complexity and scale, we reported FLOPs and the number of parameters.

The structure measure evaluates the structural similarity between a predicted saliency map $P$ and the ground truth $G$ by combining object-aware and region-aware structural similarities:
\begin{equation}
S_{m} = \alpha \cdot S_{o}(P, G) + (1-\alpha) \cdot S_{r}(P, G),
\end{equation}
where $S_{o}$ denotes object-aware similarity, $S_{r}$ denotes region-aware similarity, and $\alpha \in [0,1]$ is a weighting factor.

The mean absolute error measures the pixel-wise difference between prediction and ground truth:
\begin{equation}
M = \frac{1}{H \times W} \sum_{i=1}^{H} \sum_{j=1}^{W} \left| P(i,j) - G(i,j) \right|,
\end{equation}
where $H$ and $W$ denote the height and width of the saliency map.

The F-measure integrates precision and recall under a threshold $t$:
\begin{equation}
F_{\beta}(t) = \frac{(1+\beta^{2}) \cdot \text{Precision}(t) \cdot \text{Recall}(t)}{\beta^{2} \cdot \text{Precision}(t) + \text{Recall}(t)},
\end{equation}
where $\beta^{2}$ is typically set to $0.3$ to emphasize precision. The maximum F-measure is then defined as:
\begin{equation}
F_{m} = \max_{t \in [0,1]} F_{\beta}(t).
\end{equation}

The enhanced alignment measure evaluates both global and local alignment between $P$ and $G$:
\begin{equation}
E_{\phi}(t) = \frac{1}{H \times W} \sum_{i=1}^{H} \sum_{j=1}^{W} \phi\left(P_{t}(i,j), G(i,j)\right),
\end{equation}
where $\phi(\cdot)$ is the enhanced alignment function and $P_{t}$ is the binarized prediction under threshold $t$. The maximum enhanced alignment measure is:
\begin{equation}
E_{m} = \max_{t \in [0,1]} E_{\phi}(t).
\end{equation}

\begin{table}[t]
    \centering 
    \caption{Quantitative comparison of our HVPNet against other SOTA \textbf{RGB-T SOD} methods. Bolded results are the best.}
    \setlength{\tabcolsep}{0.3mm} 
\resizebox{\linewidth}{!}{
\begin{tabular}{c|cc|cccc|cccc|cccc}
    \toprule \multirow{2}{*}{Methods} & \multirow{2}{*}{Params(M) $\downarrow$} & \multirow{2}{*}{FLOPs(G) $\downarrow$}  & \multicolumn{4}{c|}{VT821 \cite{VT821}} & \multicolumn{4}{c|}{VT1000 \cite{VT1000}} & \multicolumn{4}{c}{VT5000 \cite{VT5000}} \\
&  &  & $E_m \uparrow$ & $S_m \uparrow$ & $F_m \uparrow$& $\mathcal{M} \downarrow$ & $E_m \uparrow$ & $S_m \uparrow$ & $F_m \uparrow$& $\mathcal{M} \downarrow$& $E_m \uparrow$ & $S_m \uparrow$ & $F_m \uparrow$  & $\mathcal{M} \downarrow$\\ \hline 
\rowcolor{yellow!20} \multicolumn{15}{c}{Lightweight Methods} \\
 OSRNet(TIM'22)~\cite{OSRNet}& 15.6 & 27.2 & .904 &\textbf{ .875} & .823& .043 & .934 & \textbf{.926} & .879 &\textbf{.022}& .915 &.876 & .828& .040 \\
LSNet(TIP'23)~\cite{lsnet}& 4.6 & 1.2 & .911 & .878 & .839 & .033& .936 & .924 & .877 & .024 & .916 &.876 & .827& .037 \\
STANet(TIP'25)~\cite{STANet1} & 5.2 & 1.5 & .915 & .878 & .827& .034& .935 & .924 & .878& .023& .918 &.880 & .831& \textbf{.035}\\
\textbf{HVPNet(-)} & 3.6 & 3.4& \textbf{.916} & .875 & \textbf{.841}& \textbf{.032} & \textbf{.946} & .918 &\textbf{.887}& .023  & \textbf{.921} & \textbf{.885} & \textbf{.833} & .036 \\

 \rowcolor{red!10} \multicolumn{15}{c}{Accurate Methods} \\

 GRNet(KBS'23)~\cite{GRNet} & 123.8 & 41.9 & .928 & .893 & .853& .031 & .961 & .931 & .908& \textbf{.018}  & .927 & .888 & .855& .034  \\
 VST-T++(TAPMI'24)~\cite{VST++}  & 100.51 & - & .912 & .894 & .846 & .037& .972 & \textbf{.941} & .931 & .020 & .933 & .895 & .854 & .034 \\
HFENet(DSP'24)~\cite{hfenet}& 52.9 & 80.1 & .909 & .881 & .837& .035& .957 & .927 & .904 & .021& .921 &.882 & .853 & .035 \\
LAFB(TCSVT'24)~\cite{LAFB}& 398.0 & 552.1 & .904 &  .883 & .824& .036  & .939 &  .928 & .897& .021 & .917 & .890 & .838 & .033 \\
SOMANet(TVC'25)~\cite{SOMANet}& 218.2 & - & .913 &  .890 & .851 & .033 & .973 & .940 & .933& .020 & .939 & .905 & .866 & .032 \\
 HVPNet &\textbf{16.6} &\textbf{11.5}& \textbf{.944} & \textbf{.907} & \textbf{.890} & \textbf{.025} & \textbf{.973} & .939 & \textbf{.931}& .019 & \textbf{.949} & \textbf{.908} & \textbf{.879} & \textbf{.030} \\
\hline
\end{tabular}
\label{tab:3}
}
\vspace{-6mm}
\end{table}
\begin{table}[t]
    \centering 
    \caption{Quantitative comparison of our HVPNet against other SOTA \textbf{RGB} and \textbf{RGB-D COD} methods. Bolded results are the best.}
    \setlength{\tabcolsep}{0.3mm} 
\resizebox{\linewidth}{!}{
\begin{tabular}{c|cc|cccc|cccc|cccc}
\toprule \multirow{2}{*}{Methods} & \multirow{2}{*}{Params(M)$\downarrow$} & \multirow{2}{*}{FLOPs(G)$\downarrow$}  & \multicolumn{4}{c|}{CAMO \cite{CAMO}} & \multicolumn{4}{c|}{COD10K \cite{COD10K}} & \multicolumn{4}{c}{NC4K \cite{NC4K}}  \\
&  & & $E_m \uparrow$ & $S_m \uparrow$ & $F_m \uparrow$& $\mathcal{M} \downarrow$& $E_m \uparrow$ & $S_m \uparrow$ & $F_m \uparrow$& $\mathcal{M} \downarrow$& $E_m \uparrow$ & $S_m \uparrow$ & $F_m \uparrow$  & $\mathcal{M} \downarrow$\\ \hline 

 \rowcolor{red!10} \multicolumn{15}{c}{Accurate Methods (RGB)} \\
 SegMar(CVPR'22)~\cite{segMar}& 56.9 & 33.4  & .884 & .816 & .803 &.071 & .907 & .833 & .755&.034 & .907 & .841 & .827  &.046\\
 FEDER(CVPR'23)~\cite{FEDER} & 44.1 & 36.2  & .873 & .802 & .789 &.071 & .905 & .822 & .768 &.032& .915 & .847 & .833  &.044\\
FSEL(ECCV'24)~\cite{FSEL}& 29.1 &35.6  & .893 & .821 & \textbf{.833} &.067& .898 & .830 & \textbf{.802}&.031 & .914 & .854 & .855&.042\\
CamoF(TAPMI'24)~\cite{camoformer}& 36.1 & 34.2  & .884 & .817 & .752&.067 & .900 &.838 & .724&.029 & .913 & .855 & .788&.042 \\
CDP(ESWA'25)~\cite{CDP}& 28.4 & - & .894 & .797 & .767&.071& .862 & .791 & .703&.036 & .902 & \textbf{.880} & .798&.050\\
 HVPNet &\textbf{11.7} &\textbf{8.1} & \textbf{.900} & \textbf{.834}& .827&\textbf{.060} & \textbf{.924} & \textbf{.841} & .790 &\textbf{.027}& \textbf{.927} &.865 & \textbf{.855}&\textbf{.038} \\
 \rowcolor{yellow!20} \multicolumn{15}{c}{Lightweight Methods (RGB-D)} \\
 LSNet(TIP'23)~\cite{lsnet} & 4.6 & 1.2& .787 & .745 & .703&.096 & .851  &.786 & .698 &.040 & .857 & .805 & .766 &.060  \\
LCOD(TVC'25)~\cite{LCOD}& 3.6 & 25.3 & .819 & .774  &.733 &.079 & .868 & .799  &.706  &.038 & .875 & .823 & .781&.050\\
 \textbf{HVPNet(-)} & 3.6 & 3.4& \textbf{.822} & \textbf{.780} & \textbf{.739}& \textbf{.076} & \textbf{.872} & \textbf{.805}& \textbf{.711}& \textbf{.037}  & \textbf{.886}& \textbf{.831}  & \textbf{.791} & \textbf{.048}\\

 \rowcolor{red!10} \multicolumn{15}{c}{Accurate Methods (RGB-D)} \\
 CIRNet(ICCV'22)~\cite{CIPNet}& 103.2 & 17.3  & .756 & .745 & .653&.100  & .832 & .806 & .669 &.040& .845 & .823 & .741 &.060 \\
PopNet(ICCV'23)~\cite{PopNet}& 198.9 & 108.0  & .769 & .719 & .657&.071  & .840 & .774 & .666&.031 & .840 & .789 & .726&.043  \\
DAINet(MS'24)~\cite{DAINet}& 47.9 &23.4  & .868 & .797 & .770 &.079& .896 & .807 & .734&.029& .905 & .839 & .812&.050\\
DSAM(ACM MM'24)~\cite{DSAM}& 118.6 & - & \textbf{.920} & .832 & .834 &.061& .931 &.846 & .807 &.033& \textbf{.940} & .871 & .862&.040 \\
 HVPNet &\textbf{16.6} &\textbf{11.5} & .919 & \textbf{.856} & \textbf{.849} &\textbf{.054} & \textbf{.935} & \textbf{.860} & \textbf{.818} &\textbf{.025}& .937 &\textbf{.880} & \textbf{.868}&\textbf{.035} \\
\hline

\end{tabular}

\label{tab:5}
}
\vspace{-5mm}
\end{table}
\begin{table}[t]
    \centering 
    \caption{Quantitative comparison of our HVPNet against other SOTA \textbf{VSOD} methods. Bolded results are the best.}
    \setlength{\tabcolsep}{0.3mm} 
\resizebox{\linewidth}{!}{
\begin{tabular}{c|c|c|cccc|cccc}
\toprule \multirow{2}{*}{Methods} & \multirow{1}{*}{Params} & \multirow{1}{*}{FLOPs}  & \multicolumn{4}{c|}{FBMS \cite{FBMS}}  & \multicolumn{4}{c}{SegV2 \cite{Segv2}}  \\
& (M) & (G) & $E_m \uparrow$ & $S_m \uparrow$ & $F_m \uparrow$& $\mathcal{M} \downarrow$& $E_m \uparrow$ & $S_m \uparrow$ & $F_m \uparrow$& $\mathcal{M} \downarrow$ \\ \hline 
 UGPL(NIPS'22)~\cite{ugpl}& - & - & .939 & .897  & .884  & .035 & .938 & .867 & .828 & .017\\
 PMN(WACV'23)~\cite{PMN}& 160.2 & 93.8 & .910 & .864  & .856  & .088& .937 & .872 & .815 & .060 \\
CFormer(TNNLS'23)~\cite{CoSTFormer}& 333.2 & 287.4 & .913 & .869  & .861  & .046& .943 & .874 & .813  & .016\\
UFO(TMM'23)~\cite{UFO}& 55.9 & - & .911 & .868  & .858  & .051 & .951 & .888 & .805 & .014 \\
SAM(TCSVT'24)~\cite{SAM(1)}& 120.0 & 101.6 & - & .872  & .842& .036 & -  & .884 & .841 & .014 \\
MMNet(TIP'24)~\cite{MMNet}& 50.8 & 82.6 & .939 &\textbf{ .897} & .884 & \textbf{.032} & .938 & .867 & .828& .014\\
 HVPNet &\textbf{16.6} &\textbf{11.5} & \textbf{.941} & .896 & \textbf{.886}& .033  & \textbf{.969} & \textbf{.917} &\textbf{.895} & .020 \\
\hline
\end{tabular}
\label{tab:4}
}
\vspace{-5mm}
\end{table}
\begin{table}[t]
    \centering 
    \caption{Quantitative comparison of our HVPNet against other SOTA \textbf{VCOD} methods. Bolded results are the best.}
\resizebox{\linewidth}{!}{
\begin{tabular}{c|c|ccccc}
\toprule \multirow{2}{*}{Methods} & \multirow{1}{*}{Params}  & \multicolumn{5}{c}{CAD \cite{CAD}} \\
& (M) & $S_a \uparrow $& $F_{\beta}^w \uparrow$& $M \downarrow$ & $mDice \uparrow$ & $mIoU \uparrow$ \\ \hline 
ZoomNet(CVPR'22)~\cite{ZoomNet}& 32.3 & .633 & .349 & .033 & .349 & .273\\
SLI-Net(CVPR'22)~\cite{SLINet}& 82.3 & .696 & .471 & .031 & .480 & .392\\
DGNet(MIR'23)~\cite{DGNet}& 21.0 & .686 & .416 & .037 & .456 & .340\\
FEDER(CVPR'23)~\cite{FEDER}& 44.2 & .691 & .444 & \textbf{.029} & .474 & .375\\
FSPNet(CVPR'23)~\cite{FSPNet} & 274.2 & .681 & .401 & .044 & .446 & .332  \\
IMEX(TMM'24)~\cite{IMEX}& 32.1 &   .695 & .490& .030 & .501 & \textbf{.412}\\
FSEL(ECCV'24)~\cite{FSEL}& 29.6 & .649 & .368 & .053 & .434 & .325\\
HGINet(TIP'24)~\cite{HGINet}& - & .680 & .437 & .050 & .501 & .392\\
TSPSAM-P(CVPR'24)~\cite{TSPSAM}& 89.7 & .681 & .500 & .031& .496 & .393\\
HVPNet &\textbf{16.6} &\textbf{.708} & \textbf{.554}& .037 & \textbf{.513} & .389 \\
\hline
\end{tabular}
\label{tab:6}
}
\end{table}
\begin{figure*}[t]
    \centering
     \includegraphics[width=\textwidth]{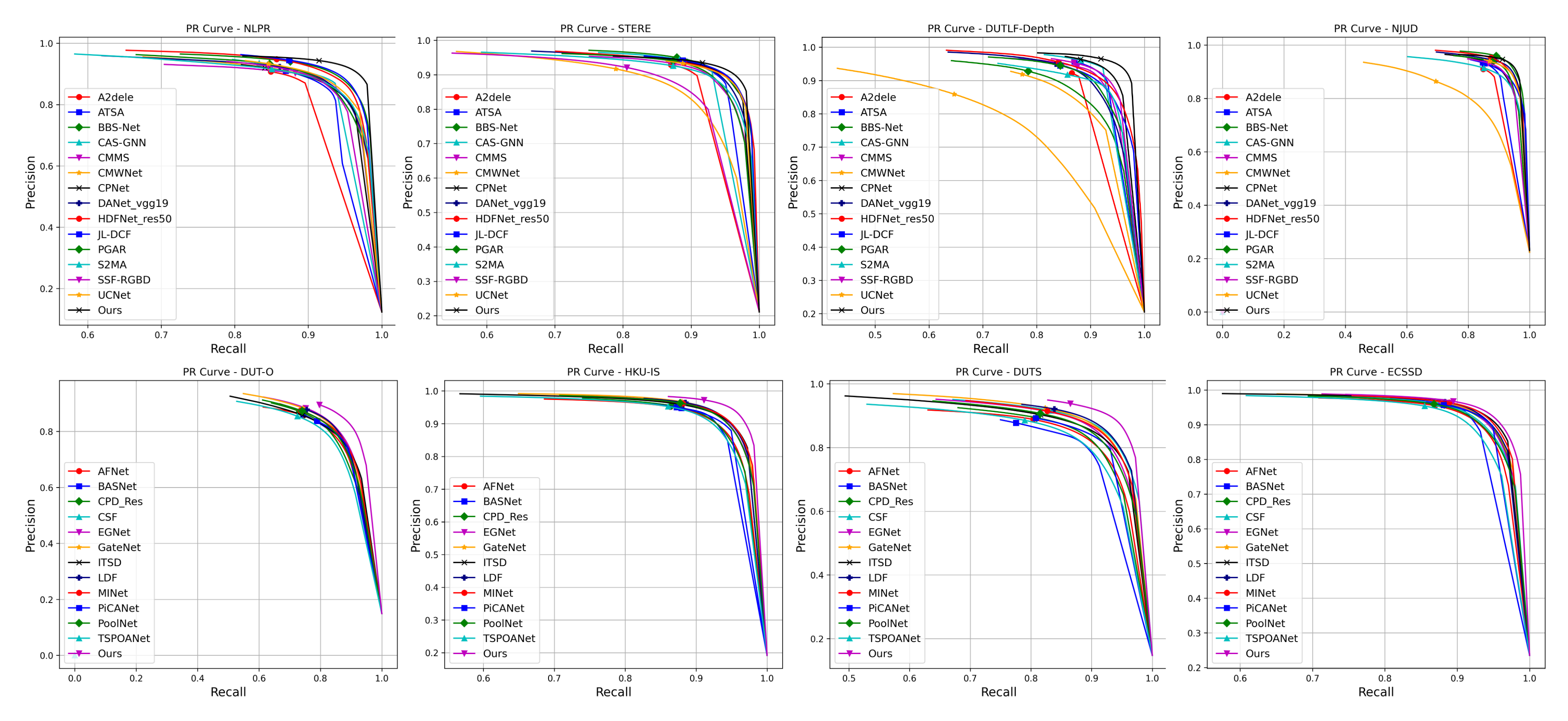}
     \vspace{-5mm}
    \caption{PR curves comparison of different models on eight RGB and RGB-D SOD datasets.}
    \label{fig:image5}
\end{figure*}
\subsection{Implementation details}
Our HVPNet model is implemented based on PyTorch and trained on an NVIDIA RTX 3090 GPU. Following previous works, we adopt the following training datasets for each task \cite{vscode,PDNet}. To ensure a fair comparison with existing methods, all images are resized to 384×384 pixels and randomly cropped to 352×352 during training. For optimization, we use the Adam optimizer with an initial learning rate of 5e-5. The learning rate is reduced by a factor of 10 after 80 epochs, and the model is trained for a total of 200 epochs until convergence. 

we employ the following datasets to train our model: the training set of DUTS~\cite{DUTS} for \textbf{RGB SOD}, the training sets of NJUD~\cite{NJU2K}, NLPR~\cite{NLPR}, and DUTLF-Depth~\cite{DUT-Depth} for \textbf{RGB-D SOD} , the training set of VT5000~\cite{VT5000} for \textbf{RGB-T SOD}, the training sets of DAVIS, FBMS and DAVSOD for \textbf{VSOD}, the training sets of COD10K~\cite{COD10K} and CAMO~\cite{CAMO} for \textbf{RGB COD}, the training sets of COD10K and CAMO for RGB-D COD, and the training set of MoCA-Mask~\cite{MoCA-Mask} for \textbf{VCOD}.

For backbone initialization, we adopt pretrained weights from SMT~\cite{smt} and MobileNetV2~\cite{MobileNetV2} to accelerate convergence and enhance overall performance. The standard HVPNet employs a stronger feature extractor to obtain more discriminative multi-level representations, thereby ensuring higher prediction accuracy.

To accommodate diverse requirements in computational efficiency and accuracy across different application scenarios, we further develop a lightweight variant, HVPNet (–), which utilizes two MobileNetV2 branches as dual backbones. 
Compared with the standard HVPNet, this design significantly reduces model complexity and computational cost, while maintaining competitive performance. By replacing the heavier backbone with lightweight MobileNetV2 encoders, HVPNet (–) achieves a more favorable trade-off between accuracy and efficiency, making it particularly suitable for resource-constrained real-world applications.

For VSOD and VCOD tasks, we follow the common practice of utilizing Flownet2.0~\cite{flownet} as the optical flow extractor due to its consistently strong performance. 

\begin{figure*}[htbp]
    \centering
     \includegraphics[width=0.8\textwidth]{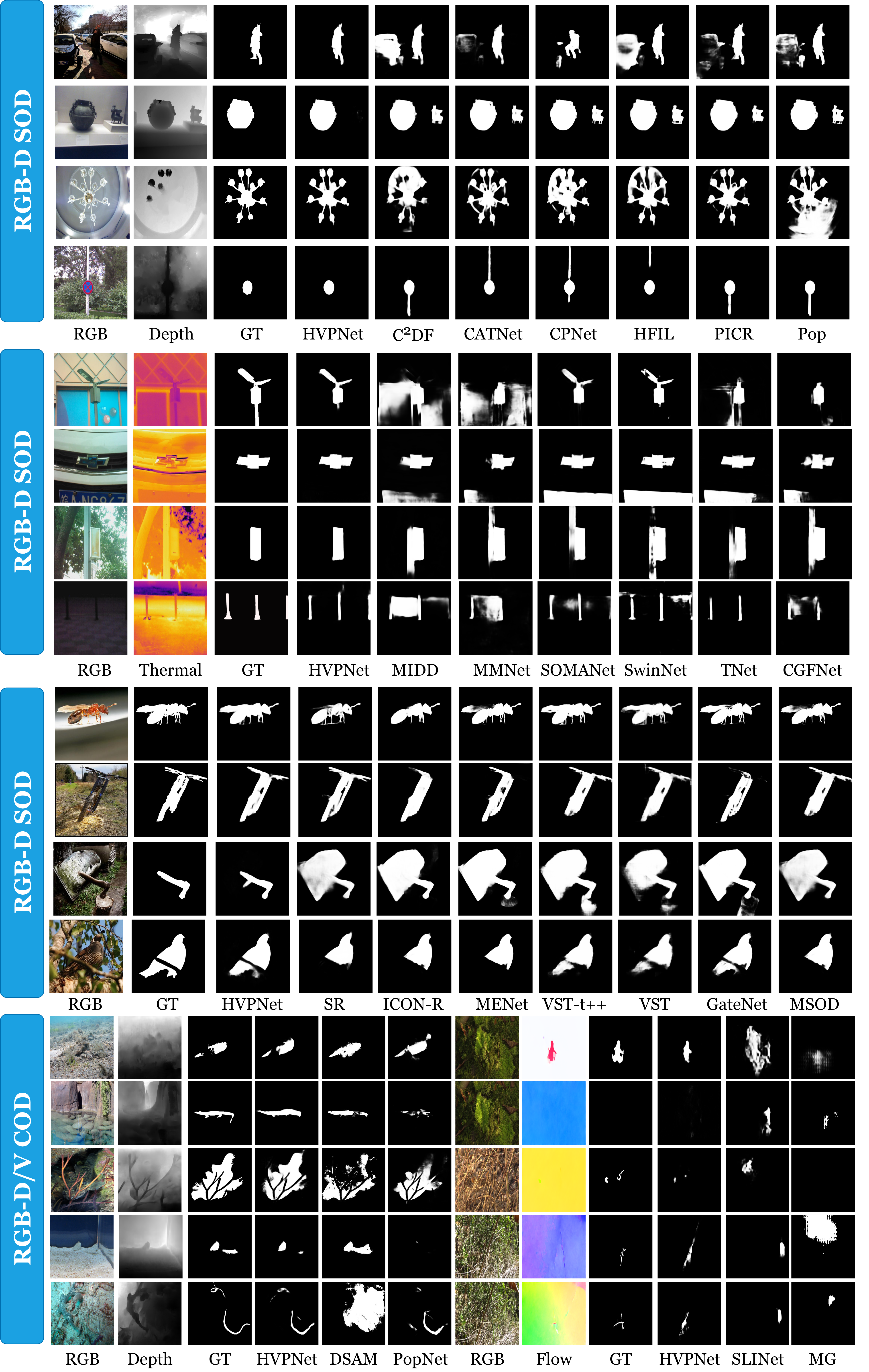}
    \caption{Qualitative comparison of our model with state-of-the-art methods, including \textbf{RGB-D SOD}, \textbf{RGB SOD}, \textbf{RGB-T SOD}, \textbf{RGB-D COD} and \textbf{VCOD}. (GT denotes ground truth.)}
    \label{fig:image6}
\end{figure*}
\subsection{Comparisons with the State-of-the-arts}
As shown in Tables \hyperref[tab:1]{\textcolor{red}{1}} through \hyperref[tab:6]{\textcolor{red}{6}}, our method demonstrates superior accuracy across all tasks, surpassing existing state-of-the-art methods. Notably, the parameters and FLOPs of HVPNet are significantly lower than those of the SOTA models. The proposed HVPNet shows consistent superiority over previous models across all seven tasks, highlighting the effectiveness of the designed retinal integration module (RIM) and cortical decoder (CD). To further validate the effectiveness of our method, the results shown in Figures \hyperref[fig:image6]{\textcolor{red}{6}} demonstrate the exceptional capabilities of HVPNet in several challenging scenarios. These scenarios include handling extremely small and large objects, multiple objects, occluded objects, and situations with uncertain boundaries, where existing methods often encounter difficulties. The experimental results show that HVPNet excels in various complex situations, particularly in handling small objects, occluded objects, and unclear boundaries, outperforming most existing models.

For further validation of the superiority of our method, as shown in Figure \hyperref[fig:image5]{\textcolor{red}{5}}, we additionally provide the PR curves on eight RGB and RGB-D datasets.
According to the results, even when compared with significantly more complex models with several times or even tens of times more parameters, our simple approach still
achieves leading performance. This indicates that model complexity is not the only path to better performance, and provides insights for future research in related tasks.

\subsection{Failure Cases}

\begin{figure}[t]
    \centering
    \includegraphics[width=0.9\columnwidth]{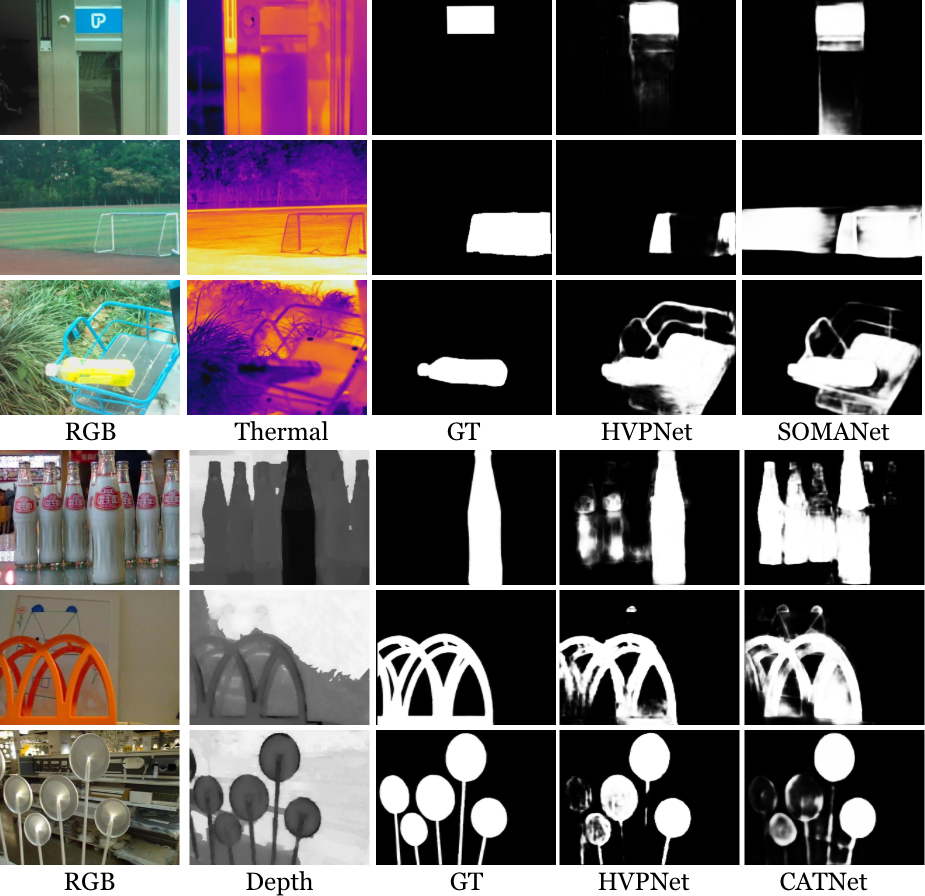}
    \vspace{-3mm}
    \caption{Illustration of failure cases. (GT: ground truth)}
    \label{fig:image7}
\end{figure}

Although HVPNet achieves competitive performance in most scenarios, it still encounters difficulties under certain extreme conditions. We further analyze these cases to better understand their underlying causes. Representative examples are shown in Figure \hyperref[fig:image7]{\textcolor{red}{7}}.
One typical failure arises when both RGB and thermal modalities consistently highlight the same irrelevant region. In such cases, the RIM-based integration strategy tends to reinforce this mutually consistent but misleading activation, ultimately producing false positives. Since the integration module is designed to exploit cross-modal consistency, it may lack sufficient capability to suppress interference when both modalities are simultaneously biased. As a result, erroneous features extracted during encoding are propagated and further amplified in the subsequent decoding stage. Another challenging scenario involves objects with transparent interiors or weak discriminative structures. When neither modality provides sufficiently distinctive cues to describe the object as a whole, the encoder struggles to learn robust and representative features. Consequently, the integrated representation lacks holistic object-level consistency, leading to incomplete or fragmented predictions. In addition, scenes with highly complex or blurred object boundaries remain difficult. The presence of intricate edges and background clutter increases the ambiguity of local features, making it challenging for the model to accurately delineate object contours while suppressing irrelevant regions. This limitation reflects insufficient boundary awareness and structural modeling under extreme conditions.

Overall, these failure cases reveal several common causes: (1) low-quality or ambiguous modal inputs that introduce unreliable features; (2) error propagation from imperfect feature extraction to integration and decoding; and (3) insufficient boundary modeling in highly complex scenes. We also observe that similar phenomena appear in other SOTA methods, suggesting that these challenges remain open problems in multimodal salient object detection.

\subsection{Ablation studies}
To verify the relative contribution of different components in HVPNet, we conducted comprehensive ablation studies by removing or replacing them within the model. These experiments are specifically designed to validate the causal relationship between the proposed bio-inspired designs and the observed performance improvements, rather than merely evaluating architectural variations.
We performed a series of experiments on four representative datasets, including one RGB-D SOD dataset, one RGB-T SOD dataset, one RGB-D COD dataset, and one VSOD dataset.
\begin{table*}[t]
    \centering 
    \caption{Ablation study of HVPNet across various task datasets. Bolded results are the best.}
    
\resizebox{\linewidth}{!}{
\begin{tabular}{c|c|c|c|ccc|ccc|ccc|ccc}
\toprule
\multirow{3}{*}{ID} & \multirow{3}{*}{Settings} & \multirow{3}{*}{Params} & \multirow{3}{*}{FLOPs} 

& \multicolumn{3}{c|}{\textbf{RGB-D SOD}} 
& \multicolumn{3}{c|}{\textbf{RGB-T SOD}} 
& \multicolumn{3}{c|}{\textbf{VSOD}} 
& \multicolumn{3}{c}{\textbf{RGB-D COD}} \\

& & & 

& \multicolumn{3}{c|}{NLPR} 
& \multicolumn{3}{c|}{VT1000} 
& \multicolumn{3}{c|}{FBMS} 
& \multicolumn{3}{c}{COD10K} \\
& &(M)&(G) 

& $E_m \uparrow$ & $S_m \uparrow$ & $F_m \uparrow$ 
& $E_m \uparrow$ & $S_m \uparrow$ & $F_m \uparrow$ 
& $E_m \uparrow$ & $S_m \uparrow$ & $F_m \uparrow$ 
& $E_m \uparrow$ & $S_m \uparrow$ & $F_m \uparrow$ \\ 
\midrule
\rowcolor{yellow!10}
& Baseline & 15.1 & 10.3 & .975  & .929 & .921 &.960 & .924 &.922 & .927 & .881 & .870  & .920 & .847 & .786 \\
\hline 
A1& MobileNetV2 + MobileNetV2 & 3.6 & 3.4 & .965  & .927 & .920 &.946 & .918 &.887 & .917 & .880 & .862  & .872 & .805 & .711 \\
 A2& SMT-t + SMT-t & 25.2 & 17.3 & .969  & .940 & .932 &.970 & .937 &.930 & .939 & .894 & .884  & .933 & .857 & .806 \\
 A3& SMT-t + ResNet18 & 25.9 & 15.8  & .970  & .941 & .932 &.971 & .938 &.930 & .939 & .895 & .883  & .932 & .858 & .815\\
 A4& PVTv2 + MobileNetV2 & 19.6 & 10.7 & .970  & .941 & .933 & 970 & .936 & .927 & .938 & .895& .884 & .933 & .857 & .815\\
A5& Swin-S + Swin-S &55.2 & 54.8 & .964  & .935 & .927 &.967 & .933 &.925 & .934 & .891 & .880  & .930 & .854 & .812\\
\rowcolor{yellow!10}
A6 & SMT-t + MobileNetV2 & 16.6 & 11.5 
& \textbf{.971} & \textbf{.942} & \textbf{.934} 
& \textbf{.973} & \textbf{.939} & \textbf{.931} 
& \textbf{.941} & \textbf{.896} & \textbf{.886} 
& \textbf{.935} & \textbf{.860} & \textbf{.818} \\
\hline 
B1&w/o RIM& 14.1 & 9.2  & .962 & .933 & .924 & .963 &.928 & .925 & .930 & .885 & .874& .924 & .850 & .806 \\
B2&stage~1& 14.3 & 10.5 & .963 & .934 & .926 & .964 &.930 & .926 & .934 & .888 & .877& .927 & .854 & .810 \\
B3&stage~1+stage~1& 15.0 & 11.2 & .967 & .937 & .930 & .969 &.936 & .929 & .938 & .891 & .880& .932 & .858 & .814 \\
B4&stage~1+stage~1+stage~1& 26.7 & 14.6 & .967 & .936 & .931 & .969 &.937 & .929 & .939 & .892 & .881& .931 & .859 & .815 \\ 
B5&stage~2+stage~2+stage~2& 17.3 & 10.5 & .968 & .936 & .930 & .970 &.937 & .928 & .937 & .892 & .881& .932 & .856 & .813\\
B6&stage~3+stage~3+stage~3& 16.2 & 10.1 & .968 & .936 & .931 & .968 &.937 & .930 & .939 & .892 & .881& .932 & .859 & .815\\
 B7& O-MMAF& 21.3 & 13.9 & .968 & .940 & .932   & .971 & .937 & .928 & .940 & .894 & .884  & .933 & .858 & .816\\
B8& CMFM & 17.1 & 11.8  & .970 & .940 & .933 & .970 & .936 & .928 & .938 & .894 & .885 & .934 & .857 & .816\\
B9 & CMA & 20.6 & 13.4& .970 & .941 & .933 & .971 &.938 & .930 & .939 & .895 & .884& .933 & .857 & .817 \\
B10 &w/o SA & 15.5 & 11.3 & .969 & .939 & .932 & .971 &.938 & .930 & .939 & .893 & .881& .934 & .859 & .817\\
\hline 
 C1& w/o Cortical Decoder& 17.5 & 12.5 & .965 & .936 & .928 & .967 &.934 & .926 & .934 & .889 & .880 & .930 & .854 & .812\\
 C2& HLDV & 16.7 & 12.0 & .967 & .938 & .930   & .971 & .937 & .929 & .938 & .893 & .884 & .933 & .858 & .817  \\
C3& LLDV& 17.4 & 12.0 & .966 & .937 & .928 & .969 &.935 & .928 & .937 & .890 & .882 & .932 & .856 & .815\\
C4&w/o GGA& 16.6 & 11.5 & .969 & .940 & .932  & .972 & .939 & .929 & .939 & .894 & .885 & .934 & .859 & .817 \\
\hline 
 D1 &w/o SRA & 15.3 & 11.1 & .968 & .939 & .932 & .970 &.937 & .929 & .938 & .894 & .883 & .933 & .858 & .816 \\
D2& {3,5,7}& 16.6 & 11.4 & .970 & .941 & .933  & \textbf{.973} & .939 & .930 & .940 & .895 & .885 & \textbf{.935} & .859 & .817 \\
D3& {5,7,9}& 16.6 & 11.4 & .970 & .940 & .932  & .972 & .938 & .929 & \textbf{.941 }& .895 & .885 & .934 & .859 & .817 \\
\rowcolor{yellow!10}
D4 & {3,5,7,9} & 16.6 & 11.5 
& \textbf{.971} & \textbf{.942} & \textbf{.934} 
& \textbf{.973} & \textbf{.939} & \textbf{.931} 
& \textbf{.941} & \textbf{.896} & \textbf{.886} 
& \textbf{.935} & \textbf{.860} & \textbf{.818} \\
\hline 
E1& $\sigma=2$& 16.6 & 11.5 & .969 & .941 & .933  & .972 & .939 & .929 & .940 & .895 & .885 & .934 & .859 & .818 \\
E2& $\sigma=6$& 16.6 & 11.5 & .970 & .940 & .932  & .972 & .939 & .930 & .939 & .894 & .885 & .934 & .859 & .817 \\
\rowcolor{yellow!10}
E3 & $\sigma=4$ & 16.6 & 11.5 
& \textbf{.971} & \textbf{.942} & \textbf{.934} 
& \textbf{.973} & \textbf{.939} & \textbf{.931} 
& \textbf{.941} & \textbf{.896} & \textbf{.886} 
& \textbf{.935} & \textbf{.860} & \textbf{.818} \\
\hline 
\end{tabular}
\label{tab:7}
}
\end{table*}
\subsubsection{Effectiveness of Different Backbone}
As shown in Table \hyperref[tab:7]{\textcolor{red}{7}}, to evaluate the impact of different backbone networks on the performance of HVPNet, we design multiple comparative experiments. In A1 we employ two MobileNetV2 networks to construct HVPNet (-). Although this configuration offers clear advantages in terms of parameters and FLOPs, the limited capability of MobileNetV2 in extracting fine-grained features
from RGB images constrains detection performance. In A2, we replace the backbones with two SMT-t networks, which leads to improved detection performance but significantly increases the parameters and FLOPs, while still failing to outperform the standard HVPNet.
Further, in A3, we utilize a CNN-based ResNet18~\cite{resnet} to extract features from the auxiliary images, while in A4, we adopt a Transformer-based PVTv2~\cite{pvtv2} to process the RGB images. Results indicate that both configurations increase the model size and computational cost, but neither achieves better detection performance than HVPNet. This suggests that using SMT-t~\cite{smt} and MobileNetV2~\cite{MobileNetV2} for feature extraction from RGB and auxiliary images, respectively, is already close to the performance ceiling. To further verify this conclusion, in A5 we adopt the more powerful Swin-S backbone for feature extraction on both branches, which not only substantially increases the parameter count and FLOPs but also leads to a notable drop in detection performance. This performance reversal demonstrates that the effectiveness of HVPNet does not originate from backbone capacity alone. Instead, it confirms that the proposed RIM and Cortical Decoder are structurally tailored to lightweight feature representations. Simply increasing backbone complexity does not yield better results, which further supports that our framework is not a trivial combination of existing modules, but a coordinated architecture optimized for efficiency–performance balance.

Crucially, as observed in our baseline settings, pairing the lightweight backbone with simple fusion and a standard decoder leads to a significant drop in accuracy, indicating that the model's performance does not rely solely on the lightweight backbone. Conversely, as shown in A5, pairing heavy backbones with the proposed modules also yields suboptimal results. This compellingly demonstrates that the proposed RIM and Cortical Decoder are specifically tailored to synergize with lightweight backbones. Our framework is not a trivial combination of existing plug-and-play modules, but a coordinated architecture optimized to achieve an excellent trade-off between accuracy and efficiency.

\subsubsection{Effectiveness of Retinal Integration Module}
As shown in Table 7, to rigorously validate the necessity and causal contribution of the bio-inspired Retinal Integration Module (RIM), we designed 10 controlled experimental settings.
 In B1, we removed the RIM and used simple addition-based fusion as the baseline. In B2 and B3, we progressively introduced stage~1 and stage~2 to assess the individual contributions of stage~1, stage~2, and stage~3. The results showed consistent performance improvement with the inclusion of each stage.
To assess the distinct roles and necessity of each RIM submodule, we conducted single Stage replacement experiments: In B4, the stage~1 replaced both stage~2 and stage~3; in B5 and B6, stage~2 and stage~3 replaced stage~1 and the other module, respectively.  Results show a marked performance drop in B4, B5, and B6. This decline occurs because stage~1 specializes in contrast enhancement and texture edge capture, stage~2 integrates structural features via the attention mechanism for mid-level understanding, and stage~3 enforces high-level semantic consistency through its optimization strategy. The three modules follow a strict processing order; interchanging any of them disrupts overall performance.

\begin{figure}[t]
	\begin{center}
			\includegraphics[width=\columnwidth]{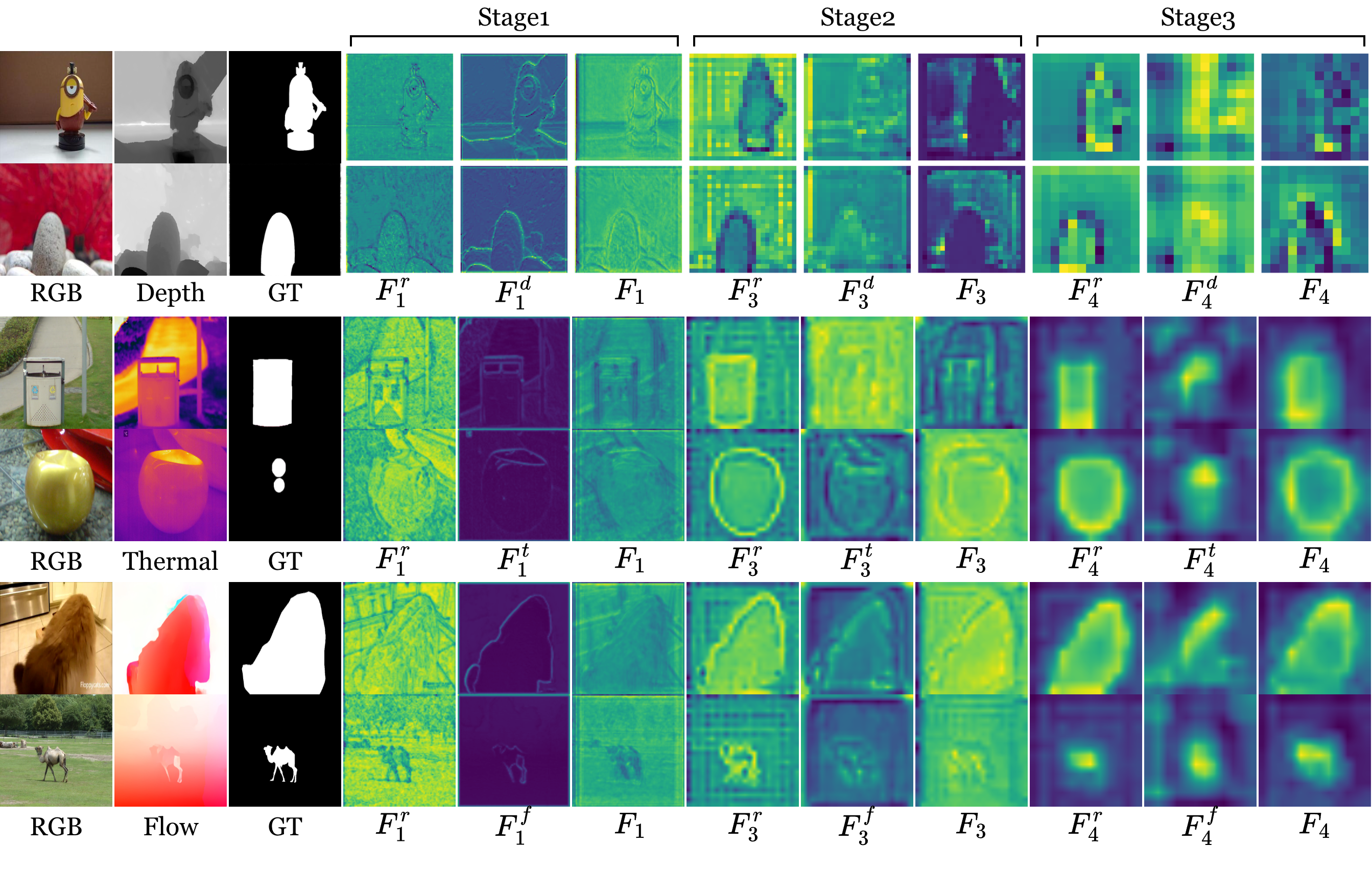}
	\end{center}
	\caption{Visualized feature maps at each stage, along with corresponding cross-modal integration results.}
	\label{fig:image8}
\end{figure}

In B7, B8, and B9, we replaced stage~1, stage~2, and stage~3 with three state-of-the-art cross-modal fusion methods \cite{SOMANet,CATNet,VST++} to compare their performance against our staged approach. Experimental results demonstrate that our method consistently outperforms these alternatives. ly, in B10, we evaluated the impact of the Spatial Attention (SA) mechanism in stage~3. The results show that SA effectively enhances semantically important regions, further improving feature representation. In addition, Figure \hyperref[fig:image8]{\textcolor{red}{8}} presents the visualized feature maps at each stage along with the corresponding cross-modal integration results, providing a more intuitive illustration of the integration process in the RIM module. The results clearly show that after integration through the RIM module, the features at each stage effectively preserve the salient target regions and contour structures.

\subsubsection{Effectiveness of Cortical Decoder}
To further validate the causal role of the bio-inspired hierarchical decoding mechanism in performance improvement, we conducted four controlled comparative experiments.
 In C1, we replaced the Cortical Decoder with a simple layer-wise additive decoding approach, which served as the baseline model. C2 integrated a High-Level Visual module into the baseline for decoding, while C3 incorporated a Low-Level Visual module. In C4, we removed the introduced GGA to evaluate its modulation effect on high-level features. In E1, E2, and E3, we further conducted an ablation study on the Gaussian kernel parameter $\sigma$. The experimental results clearly demonstrate that the model achieves the optimal performance trade-off when $\sigma=4$. The underlying reason is that an excessively small $\sigma$ results in insufficient smoothing, which fails to effectively suppress high-frequency noise and provide adequate global guidance. In contrast, an overly large $\sigma$ leads to excessive blurring of semantic features, thereby impairing accurate boundary localization. Setting $\sigma=4$ achieves an ideal balance between noise suppression and structural preservation.

We have provided more detailed data on the number of parameters and FLOPs in the supplementary materials. Experimental results showed that, compared to the baseline decoder (1.5M parameters and 2.1G FLOPs), our Cortical Decoder improved performance while reducing the number of parameters by approximately 0.9M (a 59.2\% reduction) and FLOPs by about 1.0G (a 46.7\% reduction), all while achieving better prediction accuracy. In addition, the refinement of high-level features by GGA enabled them to effectively enhance low-level visual features. These results clearly demonstrated that the proposed Cortical Decoder not only significantly improved prediction efficiency but also greatly reduced the model’s parameters and FLOPs.

\subsubsection{Effectiveness of selective region attention}
We further conducted ablation experiments within the selective region attention (SRA). Firstly, D1 removed the SRA, serving as our baseline. In D2 and D3, we used different combinations of convolution kernels: D2 employed 3×3, 5×5, and 7×7 kernels, while D3 used 5×5, 7×7, and 9×9 kernels. The experimental results showed that removing the 9×9 kernel in D2 resulted in an inability to capture finer details. Although D3 included the 9×9 kernel, it lacked the 3×3 kernel, which hindered the model’s ability to effectively attend to global information. Based on these findings, we employ 3×3, 5×5, 7×7, and 9×9 convolutional kernels as the final configuration to more effectively capture both local and global information.
\section{CONCLUSION}
In this paper, we propose HVPNet, a simple yet efficient network that draws conceptual inspiration from the human visual process for SOD and COD tasks. Moving away from the prevailing trend of architectural complexity, HVPNet abstracts the mechanisms of retinal integration and cortical hierarchical processing into level-specific integration and decoding strategies. Specifically, the proposed Retinal Integration Module (RIM) and Cortical Decoder (CD) prioritize structural compatibility between lightweight backbones and multi-level features, effectively eliminating the redundancy found in traditional "heavy fusion" paradigms. Experimental results across 7 tasks and 22 datasets demonstrate that HVPNet achieves an excellent trade-off between accuracy and efficiency, maintaining highly competitive performance with significantly fewer parameters and FLOPs. We hope this work provides a versatile baseline for multimodal object detection and encourages future research to explore the synergy between biological concepts and efficient architectural design.

\section*{Acknowledgments}
The authors declare that there are no potential conflicts of interest regarding the publication of this paper. This work is supported by grants from the National Natural Science Foundation of China (No.62262030), Jiangxi Provincial Natural Science Foundation(No.20232BAB202021).

\bigskip
\bibliographystyle{elsarticle-num} 
\bibliography{doc/IF}

@article{CATNet,
  title={CATNet: A cascaded and aggregated transformer network for RGB-D salient object detection},
  author={Sun, Fuming and Ren, Peng and Yin, Bowen and Wang, Fasheng and Li, Haojie},
  journal={IEEE Transactions on Multimedia},
  volume={26},
  pages={2249--2262},
  year={2023},
  publisher={IEEE}
}

@article{HFILNet,
  title={Heterogeneous fusion and integrity learning network for RGB-D salient object detection},
  author={Gao, Haorao and Su, Yiming and Wang, Fasheng and Li, Haojie},
  journal={ACM Transactions on Multimedia Computing, Communications and Applications},
  volume={20},
  number={7},
  pages={1--24},
  year={2024},
  publisher={ACM New York, NY}
}

@article{CPNET,
  title={Cross-modal fusion and progressive decoding network for RGB-D salient object detection},
  author={Hu, Xihang and Sun, Fuming and Sun, Jing and Wang, Fasheng and Li, Haojie},
  journal={International Journal of Computer Vision},
  volume={132},
  number={8},
  pages={3067--3085},
  year={2024},
  publisher={Springer}
}

@article{EM-Trans,
  title={Em-trans: Edge-aware multimodal transformer for rgb-d salient object detection},
  author={Chen, Geng and Wang, Qingyue and Dong, Bo and Ma, Ruitao and Liu, Nian and Fu, Huazhu and Xia, Yong},
  journal={IEEE Transactions on Neural Networks and Learning Systems},
  volume={36},
  number={2},
  pages={3175--3188},
  year={2024},
  publisher={IEEE}
}

@article{STANet1,
  title={Synergizing triple attention with depth quality for RGB-D salient object detection},
  author={Song, Peipei and Li, Wenyu and Zhong, Peiyan and Zhang, Jing and Konuisz, Piotr and Duan, Feng and Barnes, Nick},
  journal={Neurocomputing},
  volume={589},
  pages={127672},
  year={2024},
  publisher={Elsevier}
}

@article{MMNet,
  title={Disentangled cross-modal transformer for RGB-D salient object detection and beyond},
  author={Chen, Hao and Shen, Feihong and Ding, Ding and Deng, Yongjian and Li, Chao},
  journal={IEEE Transactions on Image Processing},
  year={2024},
  publisher={IEEE}
}

@INPROCEEDINGS{PopNet,
  title={Source-free depth for object pop-out},
  author={Wu, Zongwei and Paudel, Danda Pani and Fan, Deng-Ping and Wang, Jingjing and Wang, Shuo and Demonceaux, Cédric and Timofte, Radu and Van Gool, Luc},
  booktitle={ICCV}, 
  year={2023},
}

@article{CIPNet,
  title={CIR-Net: Cross-modality interaction and refinement for RGB-D salient object detection},
  author={Cong, Runmin and Lin, Qinwei and Zhang, Chen and Li, Chongyi and Cao, Xiaochun and Huang, Qingming and Zhao, Yao},
  journal={IEEE Transactions on Image Processing},
  volume={31},
  pages={6800--6815},
  year={2022},
  publisher={IEEE}
}

@inproceedings{DUT-Depth,
  title={Depth-induced multi-scale recurrent attention network for saliency detection},
  author={Piao, Yongri and Ji, Wei and Li, Jingjing and Zhang, Miao and Lu, Huchuan},
  booktitle={Proceedings of the IEEE/CVF international conference on computer vision},
  pages={7254--7263},
  year={2019}
}

@inproceedings{NJU2K,
  title={Depth saliency based on anisotropic center-surround difference},
  author={Ju, Ran and Ge, Ling and Geng, Wenjing and Ren, Tongwei and Wu, Gangshan},
  booktitle={2014 IEEE international conference on image processing (ICIP)},
  pages={1115--1119},
  year={2014},
  organization={IEEE}
}

@inproceedings{NLPR,
  title={RGBD salient object detection: A benchmark and algorithms},
  author={Peng, Houwen and Li, Bing and Xiong, Weihua and Hu, Weiming and Ji, Rongrong},
  booktitle={Computer Vision--ECCV 2014: 13th European Conference, Zurich, Switzerland, September 6-12, 2014, Proceedings, Part III 13},
  pages={92--109},
  year={2014},
  organization={Springer}
}

@inproceedings{STERE,
  title={Leveraging stereopsis for saliency analysis},
  author={Niu, Yuzhen and Geng, Yujie and Li, Xueqing and Liu, Feng},
  booktitle={2012 IEEE conference on computer vision and pattern recognition},
  pages={454--461},
  year={2012},
  organization={IEEE}
}

@article{EMAX,
  title={Enhanced-alignment measure for binary foreground map evaluation},
  author={Fan, Deng-Ping and Gong, Cheng and Cao, Yang and Ren, Bo and Cheng, Ming-Ming and Borji, Ali},
  journal={arXiv preprint arXiv:1805.10421},
  year={2018}
}

@inproceedings{SA,
  title={Structure-measure: A new way to evaluate foreground maps},
  author={Fan, Deng-Ping and Cheng, Ming-Ming and Liu, Yun and Li, Tao and Borji, Ali},
  booktitle={Proceedings of the IEEE international conference on computer vision},
  pages={4548--4557},
  year={2017}
}

@inproceedings{Fmeasure,
  title={Frequency-tuned salient region detection},
  author={Achanta, Radhakrishna and Hemami, Sheila and Estrada, Francisco and Susstrunk, Sabine},
  booktitle={2009 IEEE conference on computer vision and pattern recognition},
  pages={1597--1604},
  year={2009},
  organization={IEEE}
}

@inproceedings{Mae,
  title={Saliency filters: Contrast based filtering for salient region detection},
  author={Perazzi, Federico and Kr{\"a}henb{\"u}hl, Philipp and Pritch, Yael and Hornung, Alexander},
  booktitle={2012 IEEE conference on computer vision and pattern recognition},
  pages={733--740},
  year={2012},
  organization={IEEE}
}

@article{VT5000,
  title={RGBT salient object detection: A large-scale dataset and benchmark},
  author={Tu, Zhengzheng and Ma, Yan and Li, Zhun and Li, Chenglong and Xu, Jieming and Liu, Yongtao},
  journal={IEEE Transactions on Multimedia},
  volume={25},
  pages={4163--4176},
  year={2022},
  publisher={IEEE}
}

@inproceedings{VT821,
  title={RGB-T saliency detection benchmark: Dataset, baselines, analysis and a novel approach},
  author={Wang, Guizhao and Li, Chenglong and Ma, Yunpeng and Zheng, Aihua and Tang, Jin and Luo, Bin},
  booktitle={Image and graphics technologies and applications: 13th conference on image and graphics technologies and applications, IGTA 2018, Beijing, China, April 8--10, 2018, revised selected papers 13},
  pages={359--369},
  year={2018},
  organization={Springer}
}

@article{VT1000,
  title={RGB-T image saliency detection via collaborative graph learning},
  author={Tu, Zhengzheng and Xia, Tian and Li, Chenglong and Wang, Xiaoxiao and Ma, Yan and Tang, Jin},
  journal={IEEE Transactions on Multimedia},
  volume={22},
  number={1},
  pages={160--173},
  year={2019},
  publisher={IEEE}
}

@article{MAGNet,
  title={MAGNet: multi-scale awareness and global fusion network for RGB-D salient object detection},
  author={Zhong, Mingyu and Sun, Jing and Ren, Peng and Wang, Fasheng and Sun, Fuming},
  journal={Knowledge-Based Systems},
  volume={299},
  pages={112126},
  year={2024},
  publisher={Elsevier}
}

@article{GroupTransNet,
  title={GroupTransNet: Group transformer network for RGB-D salient object detection},
  author={Fang, Xian and Jiang, Mingfeng and Zhu, Jinchao and Shao, Xiuli and Wang, Hongpeng},
  journal={Neurocomputing},
  volume={594},
  pages={127865},
  year={2024},
  publisher={Elsevier}
}

@article{airsod,
  title={AirSOD: A lightweight network for RGB-D salient object detection},
  author={Zeng, Zhihong and Liu, Haijun and Chen, Fenglei and Tan, Xiaoheng},
  journal={IEEE Transactions on Circuits and Systems for Video Technology},
  volume={34},
  number={3},
  pages={1656--1669},
  year={2024},
  publisher={IEEE}
}

@article{OSRNet,
  title={Real-time one-stream semantic-guided refinement network for RGB-thermal salient object detection},
  author={Huo, Fushuo and Zhu, Xuegui and Zhang, Qian and Liu, Ziming and Yu, Wenchao},
  journal={IEEE Transactions on Instrumentation and Measurement},
  volume={71},
  pages={1--12},
  year={2022},
  publisher={IEEE}
}

@article{lsnet,
  title={LSNet: Lightweight spatial boosting network for detecting salient objects in RGB-thermal images},
  author={Zhou, Wujie and Zhu, Yun and Lei, Jingsheng and Yang, Rongwang and Yu, Lu},
  journal={IEEE Transactions on Image Processing},
  volume={32},
  pages={1329--1340},
  year={2023},
  publisher={IEEE}
}

@article{GRNet,
  title={Cross-modality salient object detection network with universality and anti-interference},
  author={Wen, Hongwei and Song, Kechen and Huang, Liming and Wang, Han and Yan, Yunhui},
  journal={Knowledge-Based Systems},
  volume={264},
  pages={110322},
  year={2023},
  publisher={Elsevier}
}

@InProceedings{vst,
    author    = {Liu, Nian and Zhang, Ni and Wan, Kaiyuan and Shao, Ling and Han, Junwei},
    title     = {Visual Saliency Transformer},
    booktitle = {Proceedings of the IEEE/CVF International Conference on Computer Vision (ICCV)},
    month     = {October},
    year      = {2021},
    pages     = {4722-4732}
}

@article{semanticsegmentation,
  title={Techniques and challenges of image segmentation: A review},
  author={Yu, Ying and Wang, Chunping and Fu, Qiang and Kou, Renke and Huang, Fuyu and Yang, Boxiong and Yang, Tingting and Gao, Mingliang},
  journal={Electronics},
  volume={12},
  number={5},
  pages={1199},
  year={2023},
  publisher={MDPI}
}

@article{objectdection,
  title={Object detection in 20 years: A survey},
  author={Zou, Zhengxia and Chen, Keyan and Shi, Zhenwei and Guo, Yuhong and Ye, Jieping},
  journal={Proceedings of the IEEE},
  volume={111},
  number={3},
  pages={257--276},
  year={2023},
  publisher={IEEE}
}

@misc{smt,
      title={Scale-Aware Modulation Meet Transformer}, 
      author={Weifeng Lin and Ziheng Wu and Jiayu Chen and Jun Huang and Lianwen Jin},
      year={2023},
      eprint={2307.08579},
      archivePrefix={arXiv},
      primaryClass={cs.CV}
}

@InProceedings{MobileNetV2,
author = {Sandler, Mark and Howard, Andrew and Zhu, Menglong and Zhmoginov, Andrey and Chen, Liang-Chieh},
title = {MobileNetV2: Inverted Residuals and Linear Bottlenecks},
booktitle = {The IEEE Conference on Computer Vision and Pattern Recognition (CVPR)},
month = {June},
year = {2018}
}

@article{VST++,
  title={Vst++: Efficient and stronger visual saliency transformer},
  author={Liu, Nian and Luo, Ziyang and Zhang, Ni and Han, Junwei},
  journal={IEEE Transactions on Pattern Analysis and Machine Intelligence},
  year={2024},
  publisher={IEEE}
}

@article{videosurveillance,
  title={Intersecting perspectives: Video surveillance in urban spaces through surveillance society and security state frameworks},
  author={Lysova, Tatiana},
  journal={Cities},
  volume={156},
  pages={105544},
  year={2025},
  publisher={Elsevier}
}

@article{videosummarization,
  title={Video summarization using deep neural networks: A survey},
  author={Apostolidis, Evlampios and Adamantidou, Eleni and Metsai, Alexandros I and Mezaris, Vasileios and Patras, Ioannis},
  journal={Proceedings of the IEEE},
  volume={109},
  number={11},
  pages={1838--1863},
  year={2021},
  publisher={IEEE}
}

@article{deocder1,
  title={Neural decoding of visual information across different neural recording modalities and approaches},
  author={Zhang, Yi-Jun and Yu, Zhao-Fei and Liu, Jian K and Huang, Tie-Jun},
  journal={Machine Intelligence Research},
  volume={19},
  number={5},
  pages={350--365},
  year={2022},
  publisher={Springer}
}

@article{decoder2,
  title={Leveraging the Human Ventral Visual Stream to Improve Neural Network Robustness},
  author={Shao, Zhenan and Ma, Linjian and Li, Bo and Beck, Diane M},
  journal={arXiv preprint arXiv:2405.02564},
  year={2024}
}

@InProceedings{cpd,
author = {Wu, Zhe and Su, Li and Huang, Qingming},
title = {Cascaded Partial Decoder for Fast and Accurate Salient Object Detection},
booktitle = {The IEEE Conference on Computer Vision and Pattern Recognition (CVPR)},
month = {June},
year = {2019}
}

@article{pvtv2,
  title={Pvtv2: Improved baselines with pyramid vision transformer},
  author={Wang, Wenhai and Xie, Enze and Li, Xiang and Fan, Deng-Ping and Song, Kaitao and Liang, Ding and Lu, Tong and Luo, Ping and Shao, Ling},
  journal={Computational Visual Media},
  volume={8},
  number={3},
  pages={1--10},
  year={2022},
  publisher={Springer}
}

@inproceedings{resnet,
  title={Deep residual learning for image recognition},
  author={He, Kaiming and Zhang, Xiangyu and Ren, Shaoqing and Sun, Jian},
  booktitle={Proceedings of the IEEE conference on computer vision and pattern recognition},
  pages={770--778},
  year={2016}
}

@article{CELoss,
  title={Generalized cross entropy loss for training deep neural networks with noisy labels},
  author={Zhang, Zhilu and Sabuncu, Mert},
  journal={Advances in neural information processing systems},
  volume={31},
  year={2018}
}

@InProceedings{IOU,
author = {Rezatofighi, Hamid and Tsoi, Nathan and Gwak, JunYoung and Sadeghian, Amir and Reid, Ian and Savarese, Silvio},
title = {Generalized Intersection Over Union: A Metric and a Loss for Bounding Box Regression},
booktitle = {Proceedings of the IEEE/CVF Conference on Computer Vision and Pattern Recognition (CVPR)},
month = {June},
year = {2019}
}

@article{mambasod,
  title={MambaSOD: Dual Mamba-driven cross-modal fusion network for RGB-D Salient Object Detection},
  author={Zhan, Yue and Zeng, Zhihong and Liu, Haijun and Tan, Xiaoheng and Tian, Yinli},
  journal={Neurocomputing},
  volume={631},
  pages={129718},
  year={2025},
  publisher={Elsevier}
}

@inproceedings{DUTS,
  title={Learning to detect salient objects with image-level supervision},
  author={Wang, Lijun and Lu, Huchuan and Wang, Yifan and Feng, Mengyang and Wang, Dong and Yin, Baocai and Ruan, Xiang},
  booktitle={Proceedings of the IEEE conference on computer vision and pattern recognition},
  pages={136--145},
  year={2017}
}

@inproceedings{DUT-O,
  title={Saliency detection via graph-based manifold ranking},
  author={Yang, Chuan and Zhang, Lihe and Lu, Huchuan and Ruan, Xiang and Yang, Ming-Hsuan},
  booktitle={Proceedings of the IEEE conference on computer vision and pattern recognition},
  pages={3166--3173},
  year={2013}
}

@inproceedings{ECSSD,
  title={Hierarchical saliency detection},
  author={Yan, Qiong and Xu, Li and Shi, Jianping and Jia, Jiaya},
  booktitle={Proceedings of the IEEE conference on computer vision and pattern recognition},
  pages={1155--1162},
  year={2013}
}

@inproceedings{COD10K,
  title={Camouflaged object detection},
  author={Fan, Deng-Ping and Ji, Ge-Peng and Sun, Guolei and Cheng, Ming-Ming and Shen, Jianbing and Shao, Ling},
  booktitle={Proceedings of the IEEE/CVF conference on computer vision and pattern recognition},
  pages={2777--2787},
  year={2020}
}

@inproceedings{NC4K,
  title={Simultaneously localize, segment and rank the camouflaged objects},
  author={Lv, Yunqiu and Zhang, Jing and Dai, Yuchao and Li, Aixuan and Liu, Bowen and Barnes, Nick and Fan, Deng-Ping},
  booktitle={Proceedings of the IEEE/CVF conference on computer vision and pattern recognition},
  pages={11591--11601},
  year={2021}
}

@article{CAMO,
  title={Anabranch network for camouflaged object segmentation},
  author={Le, Trung-Nghia and Nguyen, Tam V and Nie, Zhongliang and Tran, Minh-Triet and Sugimoto, Akihiro},
  journal={Computer vision and image understanding},
  volume={184},
  pages={45--56},
  year={2019},
  publisher={Elsevier}
}

@inproceedings{HKU-IS,
  title={Visual saliency based on multiscale deep features},
  author={Li, Guanbin and Yu, Yizhou},
  booktitle={Proceedings of the IEEE conference on computer vision and pattern recognition},
  pages={5455--5463},
  year={2015}
}

@inproceedings{CAD,
  title={It’s moving! a probabilistic model for causal motion segmentation in moving camera videos},
  author={Bideau, Pia and Learned-Miller, Erik},
  booktitle={Computer Vision--ECCV 2016: 14th European Conference, Amsterdam, The Netherlands, October 11-14, 2016, Proceedings, Part VIII 14},
  pages={433--449},
  year={2016},
  organization={Springer}
}

@inproceedings{MoCA-Mask,
  title={Implicit motion handling for video camouflaged object detection},
  author={Cheng, Xuelian and Xiong, Huan and Fan, Deng-Ping and Zhong, Yiran and Harandi, Mehrtash and Drummond, Tom and Ge, Zongyuan},
  booktitle={Proceedings of the IEEE/CVF Conference on Computer Vision and Pattern Recognition},
  pages={13864--13873},
  year={2022}
}

@inproceedings{FSPNet,
  title={Feature shrinkage pyramid for camouflaged object detection with transformers},
  author={Huang, Zhou and Dai, Hang and Xiang, Tian-Zhu and Wang, Shuo and Chen, Huai-Xin and Qin, Jie and Xiong, Huan},
  booktitle={Proceedings of the IEEE/CVF conference on computer vision and pattern recognition},
  pages={5557--5566},
  year={2023}
}

@inproceedings{FEDER,
  title={Camouflaged object detection with feature decomposition and edge reconstruction},
  author={He, Chunming and Li, Kai and Zhang, Yachao and Tang, Longxiang and Zhang, Yulun and Guo, Zhenhua and Li, Xiu},
  booktitle={Proceedings of the IEEE/CVF conference on computer vision and pattern recognition},
  pages={22046--22055},
  year={2023}
}

@inproceedings{vscode,
  title={Vscode: General visual salient and camouflaged object detection with 2d prompt learning},
  author={Luo, Ziyang and Liu, Nian and Zhao, Wangbo and Yang, Xuguang and Zhang, Dingwen and Fan, Deng-Ping and Khan, Fahad and Han, Junwei},
  booktitle={Proceedings of the IEEE/CVF conference on computer vision and pattern recognition},
  pages={17169--17180},
  year={2024}
}

@article{PDNet,
  title={Towards salient object detection via parallel dual-decoder network},
  author={Cen, Chaojun and Li, Fei and Li, Zhenbo and Wang, Yun},
  journal={Engineering Applications of Artificial Intelligence},
  volume={139},
  pages={109638},
  year={2025},
  publisher={Elsevier}
}

@article{SR,
  title={Towards a complete and detail-preserved salient object detection},
  author={Yun, Yi Ke and Lin, Weisi},
  journal={IEEE Transactions on Multimedia},
  volume={26},
  pages={4667--4680},
  year={2023},
  publisher={IEEE}
}

@inproceedings{DMNet-ANN,
  title={Exploring Salient Object Detection with Adder Neural Networks},
  author={Yin, Bo-Wen and Lin, Zheng},
  booktitle={Proceedings of the AAAI Conference on Artificial Intelligence},
  volume={39},
  number={9},
  pages={9490--9498},
  year={2025}
}

@article{DCNet-R,
  title={Dc-net: Divide-and-conquer for salient object detection},
  author={Zhu, Jiayi and Qin, Xuebin and Elsaddik, Abdulmotaleb},
  journal={Pattern Recognition},
  volume={157},
  pages={110903},
  year={2025},
  publisher={Elsevier}
}

@article{admnet,
  title={ADMNet: Attention-guided densely multi-scale network for lightweight salient object detection},
  author={Zhou, Xiaofei and Shen, Kunye and Liu, Zhi},
  journal={IEEE Transactions on Multimedia},
  volume={26},
  pages={10828--10841},
  year={2024},
  publisher={IEEE}
}

@article{LAFB,
  title={Learning adaptive fusion bank for multi-modal salient object detection},
  author={Wang, Kunpeng and Tu, Zhengzheng and Li, Chenglong and Zhang, Cheng and Luo, Bin},
  journal={IEEE Transactions on Circuits and Systems for Video Technology},
  volume={34},
  number={8},
  pages={7344--7358},
  year={2024},
  publisher={IEEE}
}

@article{hfenet,
  title={HFENet: Hybrid feature encoder network for detecting salient objects in RGB-thermal images},
  author={Sun, Fan and Zhou, Wujie and Yan, Weiqing and Zhang, Yulai},
  journal={Digital Signal Processing},
  volume={148},
  pages={104439},
  year={2024},
  publisher={Elsevier}
}

@article{STANet,
  title={Lightweight RGB-D Salient Object Detection from a Speed-Accuracy Tradeoff Perspective},
  author={Duan, Songsong and Yang, Xi and Wang, Nannan and Gao, Xinbo},
  journal={IEEE Transactions on Image Processing},
  year={2025},
  publisher={IEEE},
  note={Early Access}
}

@article{CoSTFormer,
  title={Learning complementary spatial--temporal transformer for video salient object detection},
  author={Liu, Nian and Nan, Kepan and Zhao, Wangbo and Yao, Xiwen and Han, Junwei},
  journal={IEEE Transactions on Neural Networks and Learning Systems},
  volume={35},
  number={8},
  pages={10663--10673},
  year={2023},
  publisher={IEEE}
}

@article{SOMANet,
  title={Semantic-Orthogonal Multi-modal Attention Network for RGB-D Salient Object Detection},
  author={Xu, Jiawei and Zhou, Qiangqiang and Yu, Jiacong and Liao, Chen and Zhu, Dandan},
  journal={The Visual Computer},
  pages={1--13},
  year={2025},
  publisher={Springer}
}

@article{LCOD,
  title={Lightweight hybrid attention RGB-D networks for accurate camouflaged object detection},
  author={Liu, Yang and Chen, Shuhan and Tang, Haonan and Wang, Shiyu},
  journal={The Visual Computer},
  pages={1--17},
  year={2025},
  publisher={Springer}
}

@article{DAINet,
  title={Depth alignment interaction network for camouflaged object detection},
  author={Bi, Hongbo and Tong, Yuyu and Zhang, Jiayuan and Zhang, Cong and Tong, Jinghui and Jin, Wei},
  journal={Multimedia Systems},
  volume={30},
  number={1},
  pages={51},
  year={2024},
  publisher={Springer}
}

@inproceedings{DSAM,
  title={Exploring deeper! segment anything model with depth perception for camouflaged object detection},
  author={Yu, Zhenni and Zhang, Xiaoqin and Zhao, Li and Bin, Yi and Xiao, Guobao},
  booktitle={Proceedings of the 32nd ACM international conference on multimedia},
  pages={4322--4330},
  year={2024}
}

@article{IMEX,
  title={Implicit-explicit motion learning for video camouflaged object detection},
  author={Hui, Wenjun and Zhu, Zhenfeng and Gu, Guanghua and Liu, Meiqin and Zhao, Yao},
  journal={IEEE Transactions on Multimedia},
  volume={26},
  pages={7188--7196},
  year={2024},
  publisher={IEEE}
}

@inproceedings{TSPSAM,
  title={Endow sam with keen eyes: Temporal-spatial prompt learning for video camouflaged object detection},
  author={Hui, Wenjun and Zhu, Zhenfeng and Zheng, Shuai and Zhao, Yao},
  booktitle={Proceedings of the IEEE/CVF conference on computer vision and pattern recognition},
  pages={19058--19067},
  year={2024}
}

@inproceedings{flownet,
  title     = {FlowNet 2.0: Evolution of Optical Flow Estimation with Deep Networks},
  author    = {Ilg, Eddy and Mayer, Nikolaus and Saikia, Tonmoy and Keuper, Margret and Dosovitskiy, Alexey and Brox, Thomas},
  booktitle = {Proceedings of the IEEE Conference on Computer Vision and Pattern Recognition (CVPR)},
  pages     = {2462--2470},
  year      = {2017}
}

@article{UFO,
title = {A Unified Transformer Framework for Group-based Segmentation: Co-Segmentation, Co-Saliency Detection and Video Salient Object Detection},
author = {Yukun Su and Jingliang Deng and Ruizhou Sun and Guosheng Lin and Qingyao Wu},
journal = {IEEE Transactions on Multimedia},
year = {2023},
publisher = {IEEE}
}

@inproceedings{PMN,
  title={Unsupervised Video Object Segmentation via Prototype Memory Network},
  author={Lee, Minhyeok and Cho, Suhwan and Lee, Seunghoon and Park, Chaewon and Lee, Sangyoun},
  booktitle={Proceedings of the IEEE/CVF Winter Conference on Applications of Computer Vision},
  pages={5924--5934},
  year={2023}
}

@inproceedings{SLINet,
  title={Implicit motion handling for video camouflaged object detection},
  author={Cheng, Xuelian and Xiong, Huan and Fan, Deng-Ping and Zhong, Yiran and Harandi, Mehrtash and Drummond, Tom and Ge, Zongyuan},
  booktitle={Proceedings of the IEEE/CVF Conference on Computer Vision and Pattern Recognition (CVPR)},
  pages={13864--13873},
  year={2022}
}

@inproceedings{ZoomNet,
  title={Zoom in and out: A mixed-scale triplet network for camouflaged object detection},
  author={Pang, Youwei and Zhao, Xiaoqi and Xiang, Tian-Zhu and Zhang, Lihe and Lu, Huchuan},
  booktitle={Proceedings of the IEEE/CVF Conference on Computer Vision and Pattern Recognition (CVPR)},
  pages={2160--2170},
  year={2022}
}

@article{FSEL,
  title={Frequency-Spatial Entanglement Learning for Camouflaged Object Detection},
  author={Sun, Yanguang and Xu, Chunyan and Yang, Jian and Xuan, Hanyu and Luo, Lei},
  booktitle={European Conference on Computer Vision},
  year={2024},
  pages={343--360},
}

@article{HGINet,
         title={Hierarchical Graph Interaction Transformer with Dynamic Token Clustering for Camouflaged Object Detection}, 
         author={Yao, Siyuan and Sun, Hao and Xiang, Tian-Zhu and Wang, Xiao and Cao, Xiaochun},
         journal={arXiv preprint arXiv:2408.15020},
         year={2024}
}

@article{CDP,
  title={Seamless Detection: Unifying Salient Object Detection and Camouflaged Object Detection},
  author={Liu, Yi and Li, Chengxin and Dong, Xiaohui and Li, Lei and Zhang, Dingwen and Xu, Shoukun and Han, Jungong},
  journal={Expert Systems with Applications},
  volume={274},
  pages={126912},
  year={2025},
  publisher={Elsevier}
}

@inproceedings{EVP,
  title={Explicit visual prompting for low-level structure segmentations},
  author={Liu, Weihuang and Shen, Xi and Pun, Chi-Man and Cun, Xiaodong},
  booktitle={Proceedings of the IEEE/CVF Conference on Computer Vision and Pattern Recognition},
  pages={19434--19445},
  year={2023}
}

@article{poolnet+,
  title={Poolnet+: Exploring the potential of pooling for salient object detection},
  author={Liu, Jiang-Jiang and Hou, Qibin and Liu, Zhi-Ang and Cheng, Ming-Ming},
  journal={IEEE Transactions on Pattern Analysis and Machine Intelligence},
  volume={45},
  number={1},
  pages={887--904},
  year={2023},
  publisher={IEEE}
}

@article{icon,
  title={Salient object detection via integrity learning},
  author={Zhuge, Mingchen and Fan, Deng-Ping and Liu, Nian and Zhang, Dingwen and Xu, Dong and Shao, Ling},
  journal={IEEE Transactions on Pattern Analysis and Machine Intelligence},
  volume={45},
  number={3},
  pages={3738--3752},
  year={2023},
  publisher={IEEE}
}

@inproceedings{MENet,
  title={Pixels, regions, and objects: Multiple enhancement for salient object detection},
  author={Wang, Yi and Wang, Ruili and Fan, Xin and Wang, Tianzhu and He, Xiangjian},
  booktitle={Proceedings of the IEEE/CVF conference on computer vision and pattern recognition},
  pages={10031--10040},
  year={2023}
}

@article{camoformer,
  title={Camoformer: Masked separable attention for camouflaged object detection},
  author={Yin, Bowen and Zhang, Xuying and Fan, Deng-Ping and Jiao, Shaohui and Cheng, Ming-Ming and Van Gool, Luc and Hou, Qibin},
  journal={IEEE Transactions on Pattern Analysis and Machine Intelligence},
  year={2024},
  publisher={IEEE}
}

@article{ugpl,
  title={Semi-supervised video salient object detection based on uncertainty-guided pseudo labels},
  author={Piao, Yongri and Lu, Chenyang and Zhang, Miao and Lu, Huchuan},
  journal={Advances in Neural Information Processing Systems},
  volume={35},
  pages={5614--5627},
  year={2022}
}

@article{FBMS,
  title={Segmentation of moving objects by long term video analysis},
  author={Ochs, Peter and Malik, Jitendra and Brox, Thomas},
  journal={IEEE transactions on pattern analysis and machine intelligence},
  volume={36},
  number={6},
  pages={1187--1200},
  year={2013},
  publisher={IEEE}
}

@inproceedings{Segv2,
  title={Video segmentation by tracking many figure-ground segments},
  author={Li, Fuxin and Kim, Taeyoung and Humayun, Ahmad and Tsai, David and Rehg, James M},
  booktitle={Proceedings of the IEEE international conference on computer vision},
  pages={2192--2199},
  year={2013}
}

@inproceedings{segMar,
  title={Segment, magnify and reiterate: Detecting camouflaged objects the hard way},
  author={Jia, Qi and Yao, Shuilian and Liu, Yu and Fan, Xin and Liu, Risheng and Luo, Zhongxuan},
  booktitle={Proceedings of the IEEE/CVF conference on computer vision and pattern recognition},
  pages={4713--4722},
  year={2022}
}

@article{DGNet,
  title={Deep gradient learning for efficient camouflaged object detection},
  author={Ji, Ge-Peng and Fan, Deng-Ping and Chou, Yu-Cheng and Dai, Dengxin and Liniger, Alexander and Van Gool, Luc},
  journal={Machine Intelligence Research},
  volume={20},
  number={1},
  pages={92--108},
  year={2023},
  publisher={Springer}
}

@article{itti,
  title={A model of saliency-based visual attention for rapid scene analysis},
  author={Itti, Laurent and Koch, Christof and Niebur, Ernst},
  journal={IEEE Transactions on Pattern Analysis and Machine Intelligence},
  volume={20},
  number={11},
  pages={1254--1259},
  year={1998},
  publisher={IEEE}
}

@inproceedings{twostream,
  title={Two-stream convolutional networks for action recognition in videos},
  author={Simonyan, Karen and Zisserman, Andrew},
  booktitle={Advances in Neural Information Processing Systems},
  pages={568--576},
  year={2014}
}

@inproceedings{salgan,
  title={Salient object detection driven by fixation prediction},
  author={Wang, Wenguan and Shen, Jianbing and Dong, Xingping and Borji, Ali},
  booktitle={Proceedings of the IEEE Conference on Computer Vision and Pattern Recognition (CVPR)},
  pages={1711--1720},
  year={2018}
}

@inproceedings{uncertainty,
  title={Uncertainty-guided transformer reasoning for camouflaged object detection},
  author={Yang, Fan and Zhai, Qiang and Li, Xin and Huang, Rui and Luo, Ao and Cheng, Hong and Fan, Deng-Ping},
  booktitle={Proceedings of the IEEE/CVF international conference on computer vision},
  pages={4146--4155},
  year={2021}
}

@inproceedings{2022recurrent,
  title={Recurrent glimpse-based decoder for detection with transformer},
  author={Chen, Zhe and Zhang, Jing and Tao, Dacheng},
  booktitle={Proceedings of the IEEE/CVF conference on computer vision and pattern recognition},
  pages={5260--5269},
  year={2022}
}

@article{DtcNet,
  title={Deep texton-coherence network for camouflaged object detection},
  author={Zhai, Wei and Cao, Yang and Xie, HaiYong and Zha, Zheng-Jun},
  journal={IEEE Transactions on Multimedia},
  volume={25},
  pages={5155--5165},
  year={2022},
  publisher={IEEE}
}

@article{exploring,
  title={Exploring figure-ground assignment mechanism in perceptual organization},
  author={Zhai, Wei and Cao, Yang and Zhang, Jing and Zha, Zheng-Jun},
  journal={Advances in Neural Information Processing Systems},
  volume={35},
  pages={17030--17042},
  year={2022}
}

@article{bicod,
  title={Bicod: a camouflaged object detection method directed by cognitive attention},
  author={Xu, Lianrui and You, Xiong and Jia, Fenli and Liu, Kangyu},
  journal={IEEE Sensors Journal},
  volume={24},
  number={4},
  pages={4711--4721},
  year={2023},
  publisher={IEEE}
}

@article{mirrornet,
  title={Mirrornet: Bio-inspired camouflaged object segmentation},
  author={Yan, Jinnan and Le, Trung-Nghia and Nguyen, Khanh-Duy and Tran, Minh-Triet and Do, Thanh-Toan and Nguyen, Tam V},
  journal={IEEE access},
  volume={9},
  pages={43290--43300},
  year={2021},
  publisher={IEEE}
}

@article{retina1,
  title={Eye smarter than scientists believed: neural computations in circuits of the retina},
  author={Gollisch, Tim and Meister, Markus},
  journal={Neuron},
  volume={65},
  number={2},
  pages={150--164},
  year={2010},
  publisher={Elsevier}
}

@article{retina2,
  title={Information processing in the primate visual system: an integrated systems perspective},
  author={Van Essen, David C and Anderson, Charles H and Felleman, Daniel J},
  journal={Science},
  volume={255},
  number={5043},
  pages={419--423},
  year={1992},
  publisher={American Association for the Advancement of Science}
}

@inproceedings{tp-seg,
  title={TP-Seg: Task-Prototype Framework for Unified Medical Lesion Segmentation},
  author={Xu, Jiawei and Zhou, Qiangqiang and Zhu, Dandan and Chen, Yong and Yi, Yugen and Zhao, Xiaoqi},
  booktitle={Proceedings of the IEEE/CVF Conference on Computer Vision and Pattern Recognition},
  pages={5452--5462},
  year={2026}
}

@article{zhou2026differseg,
  title={DifferSeg: Towards Diverse Multimodal Binary Segmentation via Differential Perception and Frequency Guidance},
  author={Zhou, Qiangqiang and Xu, Jiawei and Chen, Yong and Zhu, Dandan and Yi, Yugen and Zhao, Xiaoqi},
  journal={IEEE Transactions on Circuits and Systems for Video Technology},
  year={2026},
  publisher={IEEE}
}
\end{document}